\pdfoutput=1

\documentclass[11pt]{article}

\usepackage[preprint]{acl}

\usepackage{times}
\usepackage{latexsym}
\usepackage{booktabs}
\usepackage{multirow}
\usepackage{subcaption}  %
\usepackage{float}

\usepackage{tcolorbox}
\tcbuselibrary{skins,breakable}
\usepackage{hyperref}
\usepackage{xurl} %

\usepackage[T1]{fontenc}
\usepackage{tikz}

\usepackage[utf8]{inputenc}

\usepackage{microtype}

\usepackage{inconsolata}

\usepackage{graphicx}
\usepackage{amsmath}
\usepackage{algorithm}
\usepackage{algpseudocode} 
\usepackage[table]{xcolor}

\newcommand{\DeepFactEval}{DeepFact-Eval }

\definecolor{tradblue}{RGB}{221,235,247}   %
\definecolor{deepgreen}{RGB}{226,240,217}  %

\title{DeepFact: Co-Evolving Benchmarks and Agents for Deep Research Factuality}

\author{
  \textbf{Yukun Huang}\textsuperscript{1}\thanks{Work partially done during an internship at Amazon AGI. },
  \textbf{Leonardo F. R. Ribeiro}\textsuperscript{2},
  \textbf{Momchil Hardalov}\textsuperscript{2},
  \textbf{Bhuwan Dhingra\textsuperscript{1}} \\
  \textbf{Markus Dreyer\textsuperscript{2}},
  \textbf{Venkatesh Saligrama\textsuperscript{2,3}}
  \\
  \textsuperscript{1}Duke University,
  \textsuperscript{2}Amazon AGI,
  \textsuperscript{3}Boston University \\
  {\tt\small yukun.huang@duke.edu} 
}

\usepackage{enumitem}

\begin{document}
\maketitle
\begin{abstract}
Search-augmented LLM agents can produce deep research reports (DRRs), but verifying claim-level factuality remains challenging. Existing fact-checkers are primarily designed for general-domain, factoid-style atomic claims, and there is no benchmark to test whether such verifiers transfer to DRRs. Yet building such a benchmark is itself difficult. We first show that static expert-labeled benchmarks are brittle in this setting: in a controlled study with PhD-level specialists, unassisted experts achieve only 60.8\% accuracy on a hidden micro-gold set of verifiable claims. We propose Evolving Benchmarking via Audit-then-Score (AtS), where benchmark labels and rationales are explicitly revisable: when a verifier disagrees with the current benchmark, it must submit evidence; an auditor adjudicates the dispute; and accepted revisions update the benchmark before models are scored. Across four AtS rounds, expert micro-gold accuracy rises to 90.9\%, indicating experts are substantially more reliable as auditors than as one-shot labelers. We instantiate AtS as DeepFact-Bench, a versioned DRR factuality benchmark with auditable rationales, and DeepFact-Eval, a document-level verification agent (with a grouped lite variant) that outperforms existing verifiers on DeepFact-Bench and transfers well to external factuality datasets. \footnote{Code and data can be found in \url{https://github.com/kkkevinkkkkk/DeepFact}}
\end{abstract}

\section{Introduction}

Search-based agentic Large Language Models (LLMs) \cite{gptreserach,jin2025beneficialreasoningbehaviorsagentic} are now capable of producing deep research reports (DRRs)—complex syntheses of information that mirror expert-level analysis. These agents are increasingly deployed for scientific discovery and research, where they must synthesize vast amounts of technical literature to answer PhD-level research questions. However, verifying these complex, multi-hop scientific claims remains an open challenge. A common strategy is to check whether each claim is entailed by its cited sources \cite{deep-research-bench-Du2025DeepResearchBA, wang2025liveresearchbench}, but this ignores claims without explicit citations (often synthesized across documents) and conflates ``supported by a text'' with ``supported by scientific consensus'', ignoring cases where the cited source itself might be outdated, disputed, or cherry-picked. Reliable verification must go beyond in-report citations and cross-check the broader literature.

\begin{figure*}[t]
\centering
\includegraphics[width=0.95\linewidth]{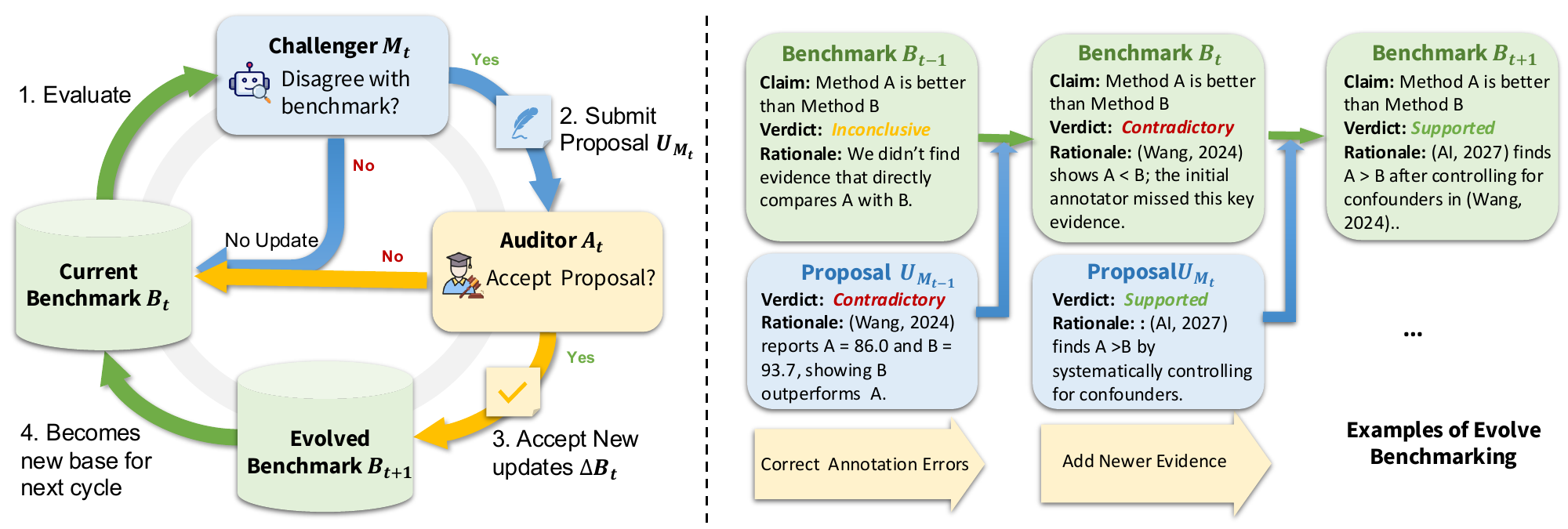}
    \caption{\textbf{Evolving Benchmarking via Audit-then-Score (AtS).} \textbf{Left:} AtS workflow. \textbf{Right:} an example of evolving benchmark. Unlike traditional static benchmarking, AtS treats ground truth  $y_i^{(t)}$ as an evolving consensus. The process proceeds in four stages: 
(1) \textbf{Evaluate:} Run a \textit{Challenger} agent ($M_t$) on the current benchmark state ($B_t$), producing a verdict $\hat{y}_i$.
(2) \textbf{Challenge:} When $\hat{y}_i \neq y_i^{(t)}$, the Challenger submits a \textit{proposal} with evidence.
(3) \textbf{Audit:} An \textit{Auditor} (human expert or trusted agent) adjudicates the dispute; if the Challenger’s argument is stronger than the incumbent rationale, the update is accepted.
(4) \textbf{Evolve \& Score:} Accepted updates yield the next benchmark state ($B_{t+1}$); the Challenger is then scored against this refined ground truth.}
    \label{fig:problems}
\label{fig:ats_protocol}
\end{figure*}

Existing automated tools \cite{wei2024longform-safe, wang-etal-2024-factcheck-gpt} that verify across web-scale sources typically focus on snippet-level matching for general-domain fact-checking. These methods, designed for simple factoids, may not be suitable for DRRs that require complex reasoning over full documents. To track future progress, the field requires an expert-level DRR factuality benchmark.

The standard benchmarking of using human experts to create a static ``gold standard'' dataset \cite{malaviya-etal-2024-expertqa, bayat2025factbench, wang-etal-2024-factcheck-gpt, thorne-etal-2018-fever}, rests on an unexamined assumption that expert judgment is infallible. Recent work \cite{xie-etal-2025-fire,nahum-etal-2025-llms-mislabel, glockner-etal-2024-ambifc-mislabel,kdd-mislabel} shows that general-domain fact-checking benchmarks contain noisy, inconsistent labels that can affect evaluations. DRR verification is harsher still. It demands deep domain expertise, reasoning over extensive context, and sustained attention, verifying a single claim can take hours, while a single report may contain hundreds of claims~\cite{patel2025deepscholarbenchlivebenchmarkautomated}. Moreover, expertise is both scarce and fragmented: even slight domain drift can make verification substantially harder, rendering multi-expert adjudication unrealistic at DRR scale.

We investigate whether experts can reliably verify DRRs factuality with a controlled study by recruiting PhD students to annotate DRR claims from their own specialties. 
Because difficult claims can take hours to adjudicate, we introduce an importance- and risk-stratified claims-sampling procedure to focus expert effort on high-impact errors. 
In parallel, we embed a hidden micro-gold set including claims that are adversarially constructed to assess annotator accuracy. We find that \emph{unassisted experts struggle even on verifiable claims within their domains} (60.8\% accuracy), suggesting that static human ``gold'' labels, and thus static benchmarks, can be unreliable for cognitively intensive expert-level reasoning tasks.

To address this, we propose \textbf{evolving benchmarking}, a new paradigm where models and benchmarks \emph{co-evolve}. We introduce the \textbf{Audit-then-Score (AtS)} protocol (see \autoref{fig:ats_protocol}) to implement this: a framework where label is not fixed but is continuously refined through auditable collaboration. When a new ``challenger'' agent's verdict disagrees with the current benchmark consensus (i.e, label and rationale), its proposal is first sent to an Auditor. The Auditor adjudicates the dispute, and the benchmark's consensus evolves if the challenger's rationale is superior. Only then are all models scored against this refined label. \textit{This process mirrors how scientific knowledge evolves—not as a frozen snapshot, but as an ongoing dialogue in which new findings can overturn prior conclusions} \cite{Elliott2017LivingSR1}.

We validate this protocol in two stages. First, we prove humans are effective auditors. We simulated the AtS process: we had experts annotate independently, then iteratively audit outputs from increasingly capable agents. Across phases, expert accuracy on hidden micro-golds improves monotonically, rising from 60.8\% to 90.9\% over four AtS rounds. This validates the AtS hypothesis: benchmark quality can co-evolve with agent capabilities, elevating the expert from a fallible labeler to a reliable auditor of a dynamic consensus. Second, we test whether the expert auditor can be replaced by an agent. Our auditor agent successfully adjudicates disputes generated by both weaker and stronger challengers, suggesting the potential for an autonomous, self-improving evaluation ecosystem.

Our validation of AtS yields two integrated artifacts. First, \textbf{DeepFact-Bench}, a DRR factuality benchmark focused on research questions. Built through multiple rounds of AtS, each claim is released with its source report, current label, and an auditable rationale that enables challenges and corrections. We will continue to evolve the benchmark after release as stronger verifiers emerge.
Second, \textbf{DeepFact-Eval}, an advanced multi-step verification agent, developed in two variants: a strong expert-level version and a lite version for efficient document-level screening, offering substantial speed and cost savings with minimal accuracy loss. 
On DeepFact-Bench, DeepFact-Eval outperforms both traditional state-of-the-art pipelines (e.g, +27.5 acc over SAFE) and repurposed deep-research verifiers (e.g, +14.3 over GPTResearcher).
DeepFact-Eval also transfers to other datasets with near-saturation performance; the remaining discrepancies appear largely due to annotation divergence, underscoring the value of benchmarks with auditable rationales and ongoing refinement.

\newcommand{\Score}{\operatorname{Score}}
\newcommand{\AtS}{\mathrm{AtS}}
\newcommand{\Static}{\mathrm{static}}
\newcommand{\MGScore}{\mathrm{MG\text{-}Score}}
\newcommand{\SCI}{\mathrm{SCI}}

\section{Related Work}
\paragraph{DRRs Evaluation} Agentic LLMs generate DRRs by iteratively retrieving and synthesizing information into long-form outputs \cite{shao2025drtulureinforcementlearning}. Current evaluations primarily focus on \emph{report-level} qualities (coherence, coverage, organization) via LLM-Judge \cite{deep-research-bench-Du2025DeepResearchBA}, expert rubrics \cite{sharma2025researchrubrics,gou2025mindweb}, experts preference \cite{chandrahasan2025deepresearchcomparatorplatform, zhao2025sciarena}, or hybrid \cite{wang2025liveresearchbench}. Factuality is typically approximated via citation checking \cite{deep-research-bench-Du2025DeepResearchBA, wang2025liveresearchbench}. However, it misses uncited claims and breaks when cited sources are biased or incomplete. We instead target global factuality, verifying against the broader scientific literature beyond the provided bibliography.

\paragraph{Fact-Checking} 
Existing fact-checking benchmarks mainly cover general-domain claims drawn from news \cite{wang-2017-liar,augenstein-etal-2019-multifc}, Wikipedia \cite{thorne-etal-2018-fever, jiang-etal-2020-hover}, LLM responses to general user instruction \cite{bayat2025factbench}, with recent work expanding into scientific domains~\cite{wadden-etal-2020-scifact, malaviya-etal-2024-expertqa}. Because they rely on human annotation, label noise from humans becomes an increasing bottleneck as models improve \cite{xie-etal-2025-fire,nahum-etal-2025-llms-mislabel, kdd-mislabel}.
Methods typically involve claim-centric workflow—claim extraction/decomposition, web retrieval, and snippet matching \cite{wei2024longform-safe, song-etal-2024-veriscore, xie-etal-2025-fire, metropolitansky-larson-2025-claimify, liu-etal-2025-verifact}. However, this shallow workflow may not transfer to DRR verification, which requires reasoning over full papers and cross-paper consistency—not only snippet-level adjudication.

\paragraph{Reliability in Human Annotations}
The reliability of human "gold" labels is increasingly contested, with judgments often compromised by cognitive biases, insufficient evidence, annotator prior, and subjectivity \cite{cognitive_biases_factcheck, atanasova-etal-2022-insufficient-fact, sap-etal-2022-annotators-prior,pavlick2019inherent}. These issues are exacerbated in high-complexity domains where expertise is fragmented. Recent benchmarks typically rely on only 1–2 experts per example \cite{malaviya-etal-2024-expertqa, openscholar}, treating them as authoritative while relying on inter-annotator agreement (IAA) as a proxy for correctness \cite{malaviya-etal-2024-expertqa,zhao2025sciarena}. However, IAA collapses unresolved disputes and fails to detect shared blind spots or systematic errors \cite{shared_blind_spot, goh2023crowdlab}. Indeed, experts are not infallible: error rates have been documented in both LLM benchmarks like HLE \cite{hle} and manual scientific literature reviews \cite{salvador-olivan2019errors}. To address these limitations in static expert annotations, we introduce adversarial micro-golds to audit annotator performance and propose a dynamic human–AI auditing framework that refines consensus over time.
See more related work in Appendix~\ref{appendix:more_related_work}.

\section{Problem Formulation}

\subsection{Task: Verifying Factuality in DRRs}
The task is to verify the factuality of claims in DRRs, whose long-form expert-level synthesis makes their verification a non-trivial reasoning task. Our goal is to assign a claim-level factuality label $y_i \in$ \{\textsc{Supported}, \textsc{Inconclusive}, \textsc{Contradictory}\} to each verifiable claim $c_i$ (see more definitions in Appendix~\ref{appendix:factuality_definition}). Each data point is a triplet $(c_i, d_i, y_i)$, where $c_i$ is a verbatim sentence and $d_i$ is the full DRR providing the context. Evaluation must consider $d_i$ to disambiguate the claim; if any part of a sentence is inconclusive or contradictory, the entire sentence inherits that label. We follow \cite{malaviya-etal-2024-expertqa} to verify at the sentence level to minimize noise from imperfect sub-claim extraction and ensure compatibility with future evaluators. Addressing the above task requires solving two coupled problems.

\paragraph{Problem 1: Modeling (Building the Verifier).}
Following prior factuality evaluation setups \cite{song-etal-2024-veriscore,wei2024longform-safe}, we focus on \emph{automated} verification to scale DRR verifications. Specifically, we build a verifier M that predicts a factuality verdict from a claim and its report context: $\hat{y}_i = M(c_i, d_i)$.

\paragraph{Problem 2: Benchmarking (Evaluating the Verifier).}
To measure the performance of any verifier $M$, we need a reliable benchmark $B$ containing ground-truth labels $y_i^{*}$ to validate verifiers' outputs. 

We first discuss Problem 2 (Benchmarking), since without a reliable benchmark $B$, verifiers (Problem 1) cannot be measured.
\subsection{Failure of Static Ground Truth} \label{sec:failure}
The traditional solution to benchmarking is the \emph{annotate-once-then-score} pipeline, which constructs a static dataset
$B={(c_i,d_i,y_i^{h})}_{i=1}^{N}$ with one-shot human labels $y_i^{h}$ \cite{wang-etal-2024-factcheck-gpt}. This paradigm implicitly treats $y_i^{h}$ as an accurate and complete proxy for the true (latent) label $y_i^{*}$, and scores a verifier $M$ by exact match:
$
\mathrm{Score}(M; S) = \frac{1}{N}\sum_i \mathbf{1}\!\left[\hat y_i = y_i^{h}\right]
$

In DRR domain, treating expert labels as a static ground truth is fragile. Even experts make mistakes in literature review~\cite{salvador-olivan2019errors}, and DRR verification is a \emph{hyper} literature-review task that requires finding and integrating evidence across sources. The long-report cognitive burden makes oversights hard to avoid, while fragmented expertise limits qualified annotators and makes multi-annotator adjudication impractical. This motivates a key question: \emph{Can experts create a valid benchmark for DRR verification?}

\section{Empirical Analysis: The Unreliability of Expert Verification}%
\label{sec:failing_paradigm}

To validate the challenges in \S\ref{sec:failure}, we ran a controlled study auditing expert reliability in realistic DRR verification. We recruited PhD-level domain specialists and used importance- and risk-stratified sampling to focus on high-stakes claims. We embedded hidden \emph{micro-gold} claims as known-answer checks to measure unassisted expert accuracy under high cognitive load. 

\subsection{Methodology: The Micro-Gold Protocol}
To measure annotation quality, we created a ``micro-gold'' set of hidden, known-answer claims, generated via a scalable, two-pronged approach required minimal domain expertise:

\paragraph{1. Unsupported Micro-Golds.} We generate unsupported micro-golds by modifying authentic DRR sentences to introduce controlled factual errors. Modifications are guided by an error taxonomy distilled from a pilot study of real model failures (detailed in \autoref{tab:error-taxonomy}), covering three cognitive stages \cite{pirolli2005sensemaking,kuhlthau1991inside}: 

\begin{itemize}[label={}, leftmargin=0pt]
    \item \textbf{Collection-stage errors:} Sourcing failures, such as hallucinated references, misattributed citations, or retrieving contextually irrelevant material. 

    \item \textbf{Analysis-stage errors:} Faulty reasoning, such as misinterpreting or incorrectly synthesizing evidence, causal inversion, or merging distinct facts into a misleading claim.
    
    \item \textbf{Generalization-stage errors:} Logical leaps, including over-generalization across domains, taxonomic simplification, or neglecting qualifiers.
\end{itemize}

Guided by this taxonomy, we injected these realistic failure modes into claims. This method is highly scalable, as we only need to confirm that our modification introduced an error, rather than performing open-ended verification. All injected errors were manually verified by authors (See examples in Appendix~\ref{appendix:adversarial_examples}).

\paragraph{2. Supported Micro-Golds.} We selected claims with explicit citations and narrow factual scope. Each candidate underwent a two-stage validation: an LLM-based entailment check against the citation, followed by a human review to confirm both the entailment and the narrow scope (e.g., verifying a specific statistic vs. a broad "SOTA" claim).

\paragraph{Usage and Validation.} These micro-golds (using a 1:4 supported-to-unsupported ratio) were hidden within annotation batches, comprising 25\% of all items. Annotator performance on this set provided a continuous measure of reliability. After the main annotation, we revealed the micro-golds to the experts, who reconfirmed their quality, further validating micro-golds (details in Appendix \ref{appendix:post_quality_check}).

\subsection{Study Setup}
To ensure annotation competence and broad domain coverage, we recruit PhD-level domain experts who are active contributors in fields such as control theory, environmental engineering, education, public health, and engineering management. Each annotator begins by proposing six research questions within their area of expertise, defined as domains in which they have at least one first-author, peer-reviewed publication. Then, among the six DRRs generated in response to these questions by deep research models (detailed in Appendix~\ref{appendix:drr_generation}), we let them choose the three they are most confident with.  
This setup ensures annotators are familiar with the subject matter, allowing them to verify complex claims accurately with minimal cognitive load. Moreover, we tell them that there are hidden tests they need to pass to get full compensation
(see Appendix~\ref{appendix:expert_annotations} for details). This overall setup ensures that they are both \emph{qualified} and \emph{well-motivated}.

\subsection{Finding 1: The 60\% Ceiling}
We had experts perform independent annotation on sampled important and risky claims (details in Appendix~\ref{appendix:imp_risk_sampling}) from DRR (we sample ~40 claims/report) in their expertise domains and score their performance on the micro-gold set.  However, they achieve only 60.8\% micro-gold accuracy—even within their specialties—showing that expert ``gold'' labels created within a finite time for complex DRR claims are unreliable. Yet multi-expert redundancy is impractical given fragmented expertise, motivating a new paradigm.

\section{Evolving Benchmark: Audit-then-Score}
\subsection{The Audit-then-Score (AtS) Protocol}
To address the susceptibility of static benchmarks $B$ to expert fallibility, we propose the \textbf{Audit-then-Score (AtS)} protocol, which maintains an \textbf{evolving benchmark} $B_t$. Rather than a one-shot ``human-gold $\rightarrow$ model evaluation'' pipeline, AtS treats benchmarking as a co-evolutionary process: model disagreements trigger auditing, and accepted revisions update the shared consensus over time. AtS involves two actors—\textbf{Challengers} (models) and \textbf{Auditors} (experts or trusted agents)—and a dynamic state, the \textbf{Consensus} ($B_t$), initialized from a seed benchmark $B_0$ built with traditional expert annotation. The loop then proceeds in two stages:

\paragraph{1. Audit:} A Challenger model $M$ is evaluated against the current benchmark consensus, $B_t$. For any claims where it disagrees, it submits a \emph{proposal} of proposed verdicts and rationales. An Auditor then adjudicates these challenges by comparing each proposed rationale against the existing one. A proposal is accepted if it demonstrates superior evidential quality or coherence, and all accepted proposals are incorporated to evolve the consensus to its next version, $B_{t+1}$.

\paragraph{2. Score:} After the benchmark is updated, the Challenger model is formally scored against the new consensus $B_{t+1}$, reflecting its performance under the refined ground truth.

In our work, the evolving benchmark maintained by AtS is instantiated as \textbf{DeepFact-Bench}. Its version at audit round $t$ is denoted \textit{DeepFact-Bench}$_t$, corresponding to benchmark state $B_t$.
Formally, AtS treats the benchmark as a versioned state $B_t$. The state at time $t+1$ is a function of the previous state, the Challenger's proposals, and the Auditor's decisions:
$
B_{t+1} = F(B_t, U_{M,t}, A_t)
$. 
Here, $B_t=\{(c_i, d_i, y_i^{(t)}, \rho_i^{(t)})\}$ is the current benchmark state, containing each claim $c_i$, its DRR context $d_i$, the current verdict $y_i^{(t)}$, and rationale $\rho_i^{(t)}$; $U_{M,t}=\{(i,\hat{y}_i,\hat{\rho}_i)\}$ denotes the set of proposals from the new Challenger $M$; and $A_t$ is the Auditor who determines whether a proposed rationale $\hat{\rho}_i$ dominates the existing one $\rho_i^{(t)}$ in evidential quality (i.e., $A_t(\hat{\rho}_i,\rho_i^{(t)})=\texttt{ACCEPT}$).

Accepted updates form the set $\Delta B_t = \{(c_i, d_i, \hat{y}_i, \hat{\rho}_i) \mid A_t(\hat{\rho}_i, \rho_i^{(t)}) = \texttt{ACCEPT}\}$,
and the benchmark then evolves to its next version:
$B_{t+1} = B_t \oplus \Delta B_t,$
where $\oplus$ denotes the update operation, producing \textit{DeepFact-Bench$_{t+1}$}.

Under AtS, each evaluated model becomes both a subject of evaluation and a potential contributor to future benchmark revisions. This transforms benchmarking from a static judgment into a continual, auditable process that co-evolves with the verifiers themselves. When later agents supply stronger reasoning or more complete evidence, the benchmark incorporates those improvements, yielding a provenance-traceable, improved ground truth. See full algorithm in Algorithm \autoref{alg:ats}.
In practice, AtS maintains versioned benchmark snapshots, changelogs, and periodic micro-gold calibration. We stop evolution when audit budget is exhausted or micro-gold performance stabilizes; all reported results are tied to a specific benchmark version. See detailed maintenance policy in Appendix \ref{appendix:manage_benchmark}.

\begin{figure}[t]
    \centering
    \includegraphics[width=0.95\linewidth]{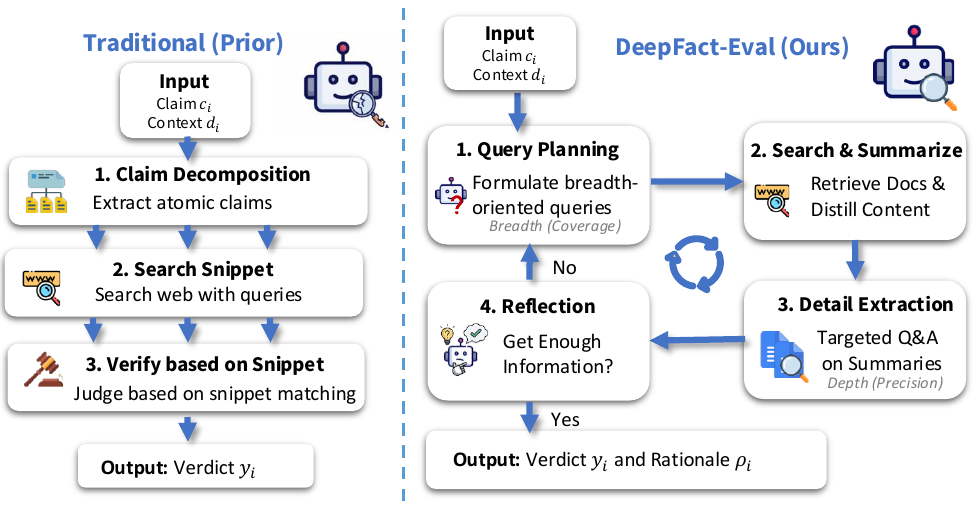}
    \caption{\textbf{DeepFact-Eval vs. traditional fact-checkers}: left, simplified VeriScore/FactCheck-GPT/SAFE; right, DeepFact-Eval workflow}
    \label{fig:deep_fact_eval}
\end{figure}

\subsection{Verification Agent: DeepFact-Eval}
\label{sec:ihar-agent}
A suite of verifier agents that serve as challengers: they are evaluated by the benchmark, yet also help build it by proposing labels and auditable rationales at scale. As they improve, they contest the current consensus and drive benchmark evolution. In our work, we adapt existing deep-research agents as verifiers and propose a stronger agentic framework, \textbf{DeepFact-Eval} (see \autoref{fig:deep_fact_eval}).

We design DeepFact-Eval to balance \emph{breadth} (document coverage) and \emph{depth} (detail precision). Given a claim and its context, the agent generates both breadth-oriented queries to retrieve relevant documents and depth-oriented questions to extract fine-grained information from them. Our methodology proceeds in the following stages: \\
\textbf{0. Claim Context Extraction:} The agent reads the whole report to extract context, unlike narrow-window approaches like VeriScore. \\
\textbf{1. Breadth-Oriented Query Planning:} It formulates diverse, targeted search queries to cover the relevant document space. \\
\textbf{2. Document Search \& Summarization:} Documents are retrieved via Google Search and summarized by an LLM to distill the content. \\
\textbf{3. Depth-Oriented Detail Questioning:} It generates follow-up questions per doc to extract claim-critical details omitted in summaries. \\
\textbf{4. Iteration or Answer}: The agent evaluates whether the collected evidence is sufficient and whether the budget remains. If evidence is insufficient and budget allows, it returns to Step 1 for another iteration; otherwise, it outputs a verdict and a rationale grounded in the retrieved documents.

To improve efficiency, we introduce \textbf{DeepFact Eval-lite}, a variant that jointly verifies semantically related claims by leveraging shared context and overlapping evidence. This reduces redundant computation while maintaining high factual rigor.

\section{Experiments: Validating AtS}
\begin{figure}[t]
    \centering
    \includegraphics[width=0.95\linewidth]{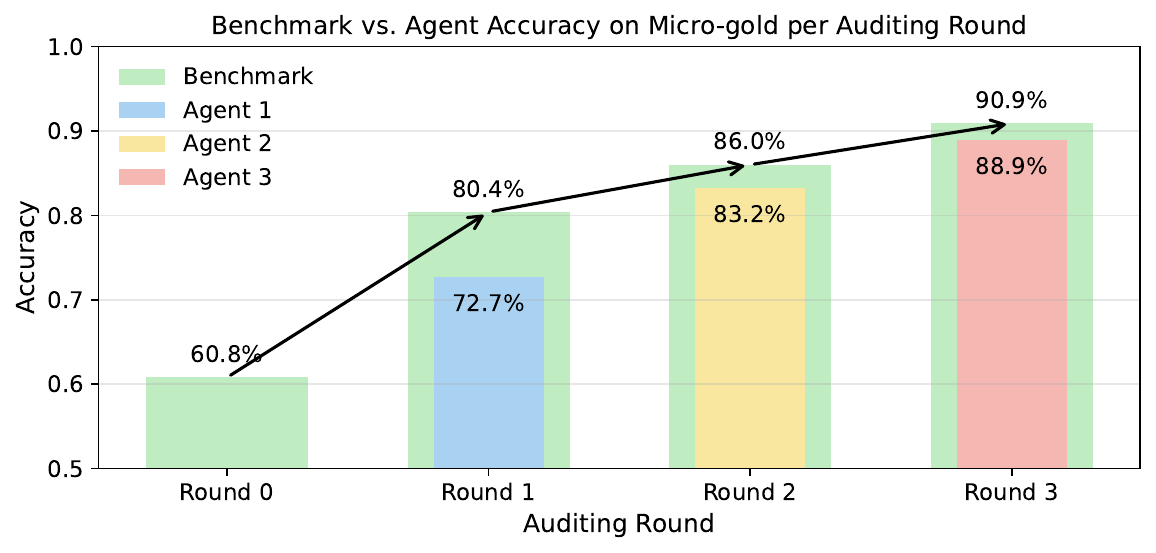}
    \caption{Benchmark Accuracy Evolution on Micro-golds Across AtS Auditing Rounds with expert auditors.}
    \label{fig:ats_result}
\end{figure}

\subsection{Simulating AtS with Human Auditors}
To study how DeepFact-Bench evolves under AtS, we simulate human experts as auditors while introducing progressively stronger agents; each round produces a new benchmark version reflecting the updated consensus. \\ \\
\textbf{Round 0 (Expert-only):}
Experts annotate claims independently to initialize the benchmark.\\ \\
\textbf{Rounds 1--3 (Expert auditing agents):}
Experts audit Agent$_i$ in round $i$ (conditioned on the previous round's consensus): 
\textbf{Agent1 (A1): SmolAgents (GPT-4.1, see \S~\ref{sec:baselines})}, \textbf{Agent2 (A2): DeepFact-Eval (GPT-4.1)}, and \textbf{Agent3 (A3): DeepFact-Eval (GPT-5)}. 
In each round, experts accept an agent's proposed revision only when its rationale provides stronger evidence or reasoning, and the updated consensus becomes the next version of \textsc{DeepFact-Bench}. We track benchmark refinement using micro-gold accuracy across these versions.

\subsection{Finding 2: Humans Are Effective Auditors}
\paragraph{Experts are fallible alone, but improve with auditing.}
Experts struggle to verify DRR claims in isolation: in \textit{Round 0} of \autoref{fig:ats_result}, micro-gold accuracy is only 60.8\% despite high self-reported confidence, highlighting how long-context, cross-source DRR verification can hide experts’ blind spots.
In contrast, auditing agent verdicts and rationales substantially improves expert accuracy, which increases monotonically as the challenger strengthens (Agent1 $\rightarrow$ Agent2 $\rightarrow$ Agent3; \autoref{fig:ats_result}), supporting AtS: fallible experts can refine benchmarks when scaffolded by strong verifiers.

\paragraph{Humans Auditor Complements Agent Challenger}
To see how agent challengers affect humans auditors, we analyze human--agent decision flows, yielding eight possible correctness transitions (\autoref{tab:decision-flow}). Frequent 011 and 101 patterns show auditors actively evaluate agents, adopting correct input and rejecting incorrect suggestions, supporting \textit{AtS} as an evidence-driven audit process.

\definecolor{darkgreen}{RGB}{0,100,0}
\definecolor{darkred}{RGB}{139,0,0}

\begin{table}[t]
\centering
\tiny
\begin{tabular}{@{}p{2cm}p{3.5cm}p{1cm}@{}}
\toprule
\textbf{Decision Flow} & \textbf{Interpretation} & \textbf{Proportion} \\
\midrule
\textcolor{darkgreen}{H:0 → A:1 → H':1} & Human learns from correct agent & 22.4\% \\
\textcolor{darkgreen}{H:0 → A:0 → H':1} & Human learns from wrong agent & 0.0\% \\
\textcolor{darkgreen}{H:1 → A:1 → H':1} & Both correct throughout & 44.8\% \\
\textcolor{darkgreen}{H:1 → A:0 → H':1} & Human resists being misled & 13.3\% \\
\midrule
\textcolor{darkred}{H:0 → A:1 → H':0} & Stays wrong despite right agent & 5.6\% \\
\textcolor{darkred}{H:0 → A:0 → H':0} & Both wrong; no improvement & 11.2\% \\
\textcolor{darkred}{H:1 → A:0 → H':0} & Human misled by wrong agent & 2.8\% \\
\textcolor{darkred}{H:1 → A:1 → H':0} & Human flips to wrong despite both correct & 0.0\% \\
\bottomrule
\end{tabular}
\caption{Human--agent decision flows on micro-gold claims, showing correctness transitions (1/0) for Human (H), Agent (A), and post-audit Human (H'). Flow patterns ending correct are in \textcolor{darkgreen}{green}; wrong ones are in \textcolor{darkred}{red}.}
\label{tab:decision-flow}
\end{table}

\begin{figure*}[h]
    \centering
    \includegraphics[width=0.95\linewidth]{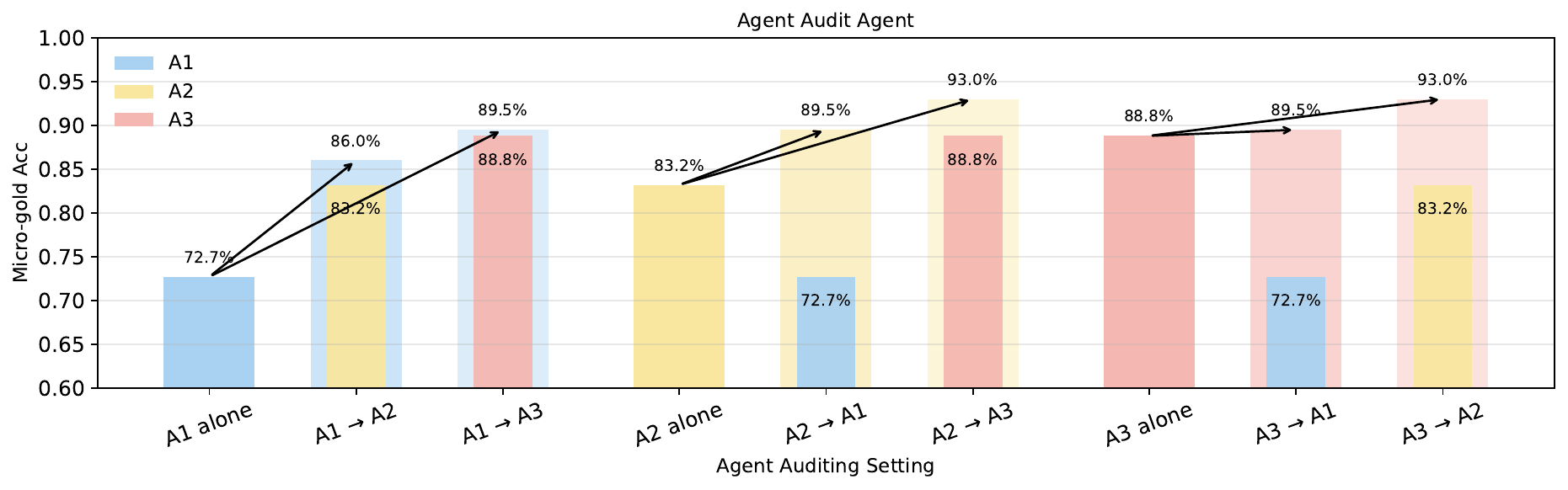}
    \caption{\textbf{Agent-only auditing for AtS}. For each auditor $A_i$, we report its Round-0 solo accuracy and its Round-1 audited accuracy when auditing another agent $A_j$ ($A_i\!\rightarrow\!A_j$; outer bars). Inner bars within each $A_i\!\rightarrow\!A_j$ show the audited agent’s solo (Round-0) accuracy $A_j$ for reference.}
    \label{fig:agent_auditor}
\end{figure*}

\subsection{Finding 3: Agents Are Auditors Proxies} 
We test whether agents can be auditors by replicating AtS with agent auditors. Round~0: each agent $A_i$ verifies claims independently. Round~1: $A_i$ audits another agent $A_j$ by adjudicating between their Round-0 outputs to produce an updated decision. For each $A_i$, we report solo accuracy and audited accuracy when auditing the other two agents. 

\paragraph{Agents are non-regressive auditors.}
As shown in ~\autoref{fig:agent_auditor}, across all pairings, auditing outperforms the audited agent’s solo micro-gold baseline in both weaker$\rightarrow$stronger and stronger$\rightarrow$weaker directions (e.g., $A2\rightarrow A3$ and $A3\rightarrow A2$ exceed both $A2$ solo and $A3$ solo), indicating that agent auditor can combine complementary evidence and catch oversights, creating a benchmark that surpasses the individual verifiers.

\paragraph{Auditing consolidates; Verifiers expand.}  Stronger$\rightarrow$weaker and weaker$\rightarrow$stronger auditing perform similarly (e.g., $A1\!\rightarrow\!A3 \approx A3\!\rightarrow\!A1$), indicating that auditing is constrained by available evidence rather than adjudication skill. Therefore, the auditor serves to \textbf{consolidate} a rigorous baseline from existing outputs, but benchmark evolution depends entirely on stronger verifiers to \textbf{expand} the information scope.

\newcommand{\colorhighlight}[2]{%
  {\setlength{\fboxsep}{1pt}%
   \colorbox{#1}{\strut\hspace{0.15em}\text{#2}\hspace{0.15em}}}%
}

\begin{table*}[htbp]
\centering
\small
\begin{tabular}{llccccccc}
\toprule
\multirow{2}{*}{Model} & \multirow{2}{*}{Backbone} & \multicolumn{4}{c}{Quality} & \multicolumn{3}{c}{Efficiency} \\
\cmidrule(lr){3-6} \cmidrule(lr){7-9}
 & & Acc & F1 & Precision & Recall & Input Tokens & Output Tokens & Cost \\
\midrule
\multicolumn{9}{c}{\textit{Main Results}} \\
\midrule
\cellcolor{tradblue}Factcheck-GPT & GPT-4.1 & 55.0 & 58.3 & 67.7 & 51.2 & -- & -- & -- \\
\cellcolor{tradblue}SAFE & GPT-4.1 & 55.9 & 53.0 & 76.3 & 40.6 & -- & -- & -- \\
\cellcolor{tradblue}VeriScore & GPT-4.1 & 52.5 & 48.9 & 71.9 & 37.0 & -- & -- & -- \\
\cellcolor{tradblue}Fire & GPT-4.1 & 58.5 & 63.2 & 69.2 & 58.3 & -- & -- & -- \\
\cellcolor{deepgreen}GPT-Researcher (Deep)  & GPT-4.1 & 69.1 & 79.7 & 66.7 & 98.9 & 52.3K & 9.0K & \$0.18 \\
\cellcolor{deepgreen}GPT-Researcher (Deep+) & GPT-4.1 & 68.3 & 79.3 & 66.1 & \textbf{99.2} & 83.3K & 13.9K & \$0.28 \\
\cellcolor{deepgreen}SmolAgents & GPT-4.1 & 68.8 & 69.5 & 58.0 & 86.7 & 294.4K & 3.4K & \$0.62 \\
\cellcolor{deepgreen}DeepFact-Eval & GPT-4.1 & \textbf{83.4} & \textbf{86.9} & \textbf{85.7} & 88.2 & 516.9K & 18.6K & \$1.16 \\
\cellcolor{deepgreen}DeepFact-Eval (Group=5) & GPT-4.1 & 77.9 & 83.1 & 78.5 & 88.2 & 131.4K & 4.9K & \$0.30 \\
\cellcolor{deepgreen}DeepFact-Eval (Group=10) & GPT-4.1 & 76.3 & 82.2 & 76.4 & 89.0 & 93.5K & 3.5K & \$0.21 \\
\midrule
\multicolumn{9}{c}{\textit{Model Ablations}} \\
\midrule
\cellcolor{deepgreen}DeepFact-Eval & GPT-5 & 87.2 & 89.9 & 87.9 & 91.9 & -- & -- & -- \\
\cellcolor{deepgreen}DeepFact-Eval & Gemini-2.5-Pro & 81.5 & 85.0 & 84.6 & 85.3 & -- & -- & -- \\
\cellcolor{deepgreen}DeepFact-Eval & Qwen-3-32B & 72.5 & 77.4 & 78.1 & 76.6 & -- & -- & -- \\
\bottomrule
\end{tabular}
\caption{\textbf{Comparison of fact-checkers on DeepFact-Bench} (accuracy/F1/precision/recall) and efficiency. Best GPT-4.1-backbone results are \textbf{bolded}. \colorhighlight{tradblue}{Traditional} and \colorhighlight{deepgreen}{deep-research} methods are color-coded.}
\label{tab:factcheck_results}
\end{table*}

\subsection{Ablations on Evolution Dynamics}
We study two design choices in benchmark evolution: \emph{how often} conflicts are audited, and \emph{how strict} the revision rule should be.

\paragraph{Audit frequency.}
Using an offline counterfactual replay, we audit a random fraction \(p\in\{0.25, 0.5, 0.75, 1.0\}\) of detected conflicts in each round; unaudited conflicts leave the benchmark unchanged. Accuracy improves across rounds for all \(p\), but higher \(p\) improves faster early: in Round-1, accuracy is \(66.4/72.0/73.4/80.4\), and by Round-2 it is \(68.5/76.2/81.8/85.3\), for \(p\)=\(0.25/0.5/0.75/1.0\). By Round-3, performance converges at higher sampling rates: \(76.2/85.3/89.5/90.9\), with \(p=0.75\) within \(1.4\) points of full auditing. This suggests that audit frequency mainly affects the speed of early improvement, while later rounds show diminishing returns.

\paragraph{Revision strictness.}
We also test a stricter revision policy that applies updates only when both the human auditor and an agent auditor agree. Its effect is mixed: it improves Round-2 micro-gold accuracy from \(0.86\) to \(0.88\), but slightly reduces Round-3 accuracy from \(0.909\) to \(0.902\). This suggests that stricter gating can filter noisy revisions but also block beneficial updates in others.

Overall, these results show that benchmark evolution involves a practical tradeoff between cost, speed, and conservativeness, and that AtS supports flexible maintenance policies depending on annotation budget and quality goals.

\subsection{Artifact: DeepFact-Bench}
DeepFact-Bench is our claim-level factuality benchmark for DRRs, built from expert-audited claims and maintained as an evolving benchmark under AtS after one initial round and three audit rounds, and each instance includes a verbatim claim sentence, its source report as context, a final expert verdict, and a supporting rationale. The benchmark contains 944 claims from 20 reports spanning six domains. We use 323 claims from 5 CS reports as the validation split, and 621 claims from 15 reports across the other five domains as the test split. Within the test set, 143 claims are micro-golds, of which 120 are adversarially constructed. Excluding these adversarial examples, 27.0\% of the remaining test claims are naturally unsupported. Appendix~\ref{appendix:post_quality_check} provides a post-hoc quality check, and Appendix~\ref{appendix:deep_research_erros_examples} presents qualitative examples of factual errors in DRRs.

\paragraph{Future Updates}
As verifiers improve and evidence evolves, consensus can drift. We will update the benchmark mainly with agent-auditors and use experts for periodic calibration  until micro-gold accuracy reaches 100\%.  To control cost, we trigger audits only when a challenger verifier outperforms the current benchmark on the micro-gold set, and we batch agent-audited revisions for release. Because iterative updates can subtly tune the benchmark to the reasoning style or retrieval habits of the challenger and the auditors, we additionally schedule an expert re-audit once cumulative updates exceed 5\% of benchmark verdicts to re-validate consensus and correct drift.

\paragraph{Cost and Practicality}
Constructing DeepFact-Bench required over 400 expert-hours in total, including both external experts and in-house annotators. However, AtS is not the main cost driver: the dominant cost is the one-time expert-level verification required by this inherently difficult DRR claim-checking task, which any expert-quality benchmark must pay at least once. We decompose total effort into a one-time base cost for constructing the seed benchmark \(B_0\), and the incremental AtS cost from later audit rounds. In practice, the base cost dominates, while later rounds become cheaper as conflicts shrink and experts gain familiarity with the reports, claims, and evidence. On the test set, claims requiring expert work drop from \(621\) in Round-0 to \(361\), \(247\), and \(182\) in the next three rounds, with corresponding expert time shares of \(65.5\%\), \(21.1\%\), \(7.71\%\), and \(5.68\%\). Thus, Round-0 accounts for \(65.5\%\) of total expert effort, while the three AtS rounds together account for only \(34.5\%\), yet improve benchmark accuracy from \(60.8\%\) to \(90.9\%\). Overall, AtS does not multiply annotation cost; it amortizes the unavoidable cost of expert verification into progressively cheaper follow-up rounds.

\begin{figure*}[t]
    \centering
    \includegraphics[width=0.95\linewidth]{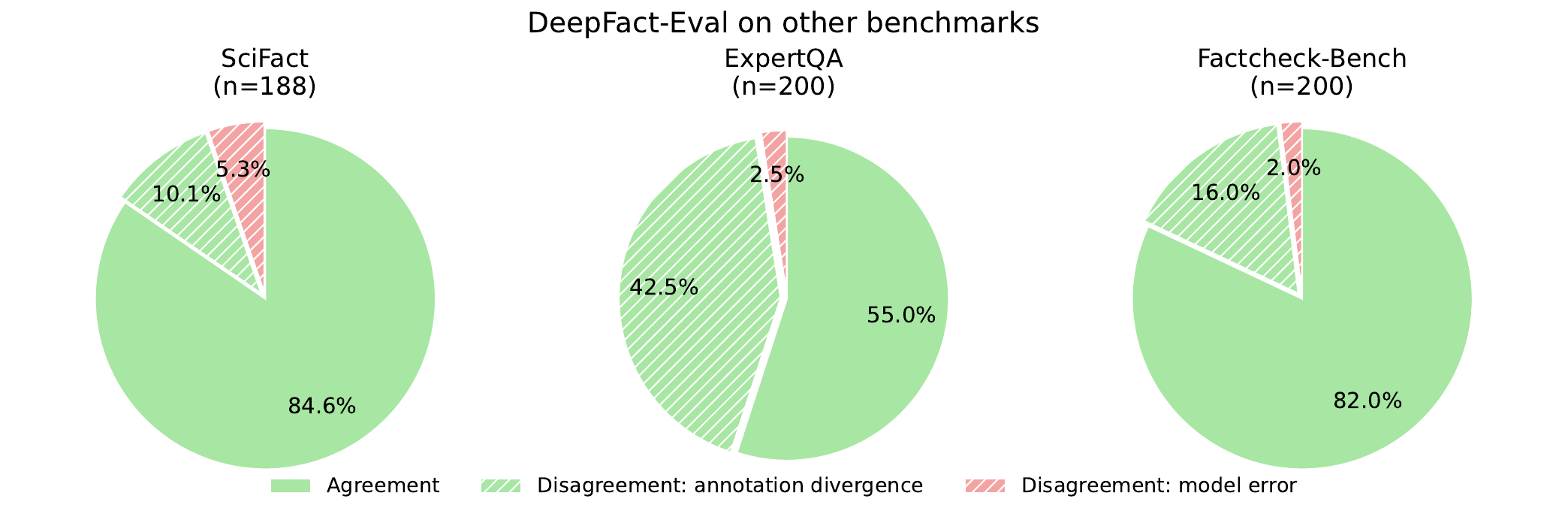}
    \caption{\textbf{Results of DeepFact-Eval on SciFact, ExpertQA, Factcheck-Bench.} Solid green indicates \emph{Agreement} (verifier's prediction matches the benchmark label). Hatched slices denote \emph{Disagreements} (verifier's prediction doesn't match the benchmark label). Green-hatched indicates \emph{Annotation divergence} (e.g., evidence--label misalignment, non-verifiable/ambiguous sentences, subjective or underspecified claims, or annotation divergence can't be resolved due to the lack of a gold rationale), while red-hatched indicates \emph{Likely model error} (expert re-annotation aligns with the benchmark label).}
    \label{fig:other_datasets}
\end{figure*}

\section{Results on DeepFact-Bench}
\subsection{Baselines}
\label{sec:baselines}
We benchmark verifiers on DeepFact-Bench test set.
\textbf{FactCheck-GPT}~\cite{wang-etal-2024-factcheck-gpt} extracts atomic claims, retrieves evidence, judges stance, and issues corrections.
\textbf{SAFE}~\cite{wei2024longform-safe} breaks responses into atomic facts and iteratively issues Google Search queries, judging support from retrieved snippets.
\textbf{VeriScore}~\cite{song-etal-2024-veriscore} follows a similar retrieve--judge paradigm but verifies only \emph{verifiable} claims in a single optimized pass.
\textbf{FIRE}~\cite{xie-etal-2025-fire} casts verification as an agentic loop: it either returns a verdict or generates a follow-up query and repeats until confident.
Finally, we repurpose deep-research-style scaffolding as a verifier baseline for comparison: \textbf{GPTResearcher (Deep Research Mode)}~\cite{gptr}, a workflow agent that iteratively performs query planning and retrieval-augmented synthesis with a tunable search-depth budget (``Deep+'' uses a larger budget), and \textbf{SmolAgents}~\cite{smolagents}, a ReAct-style agent where a main agent can invoke a search sub-agent to interact with websites and gather evidence before answering.
See Appendix~\ref{appendx:implementation} for details.

\subsection{Results}
Following \cite{song-etal-2024-veriscore}, we merge contradictory and inconclusive cases into \texttt{Unsupported}, yielding a binary supported/unsupported setting. We report accuracy, F1, precision/recall for the supported class, plus efficiency (I/O tokens and estimated cost per claim) for deep-research methods only, since snippet-based pipelines have negligible cost (details in Appendix~\ref{appendx:implementation}). As no baseline outperforms DeepFact-Eval on micro-gold, we do not run benchmark evolution and instead evaluate all methods on the current snapshot (see \autoref{tab:factcheck_results}).

\paragraph{\DeepFactEval achieves the best performance.}
\DeepFactEval attains the highest accuracy (83.4\%; \autoref{tab:factcheck_results}), outperforming both traditional fact-checking pipelines (best: 58.5\%) and prior deep-research agent baselines (best: 69.1\%). Deep-research verifiers generally outperform snippet-based methods (e.g., GPT-Researcher 69.1 vs.\ VeriScore 52.5) because DRR claims rarely have a single verbatim supporting span; evidence is often distributed across a document. This gap is reflected in the error profiles: snippet-based checkers are high-precision but low-recall and often default to \texttt{Unsupported} when retrieval fails, while general deep-research agents improve recall but sacrifice precision by accepting fuzzy topical support. \DeepFactEval closes this gap via detail verification with targeted deep queries that surface the technical distinctions that determine support, yielding both high precision and high recall. See qualitative examples of DeepFact-Eval in Appendix \ref{appendix:deepfact-eval_examples}.

\paragraph{Grouping reduces cost with minimal quality trade-off.}
\DeepFactEval reduces verification cost substantially via grouped verification, with only a minor loss in accuracy, making it a cost-efficient option. Notably, \DeepFactEval (Group=10) outperforms GPT-Researcher at a comparable budget (76.4 vs. 69.1). In contrast, scaling GPT-Researcher’s compute by increasing max search depth does not improve performance (69.1 $\rightarrow$ 68.3) despite higher claim cost (\$0.18 $\rightarrow$ \$0.28).

\paragraph{DeepFact-Eval generalizes across backbones.}
Upgrading the backbone of DeepFact-Eval to a stronger model (GPT-5) improves performance, while Gemini-2.5-Pro performs comparably to GPT-4.1. Using an open-source backbone (Qwen3-32B) reduces accuracy, but \DeepFactEval still outperforms other baselines using GPT-4.1.

\subsection{Results on Other Factuality Benchmarks}
We further test whether \textsc{DeepFact-Eval} generalizes beyond \textsc{DeepFact-Bench} by evaluating it on SciFact~\cite{wadden-etal-2020-scifact}, ExpertQA~\cite{malaviya-etal-2024-expertqa}, and Factcheck-Bench~\cite{wang-etal-2024-factcheck-gpt}. To better understand its apparent errors, we audit disagreement cases to distinguish verifier failures from benchmark artifacts. On SciFact, where evidence rationales are available, we directly compare disagreements against the provided evidence; on ExpertQA and Factcheck-Bench, where rationale support is weaker or absent, we use blinded re-annotation on disagreement subsets. Full dataset setup, preprocessing, and audit details are deferred to Appendix~\ref{appendix:other_datasets}.

We find that \textsc{DeepFact-Eval} transfers well to these external benchmarks, and that many of its residual disagreements are better explained by benchmark brittleness than by verifier failure. After auditing disagreement cases, its estimated accuracy rises from 84.6\% to 94.7\% on SciFact and from 82.0\% to 93.0\% on Factcheck-Bench. On ExpertQA, expert re-annotation also frequently sides with \textsc{DeepFact-Eval}, though the lack of gold rationales makes adjudication less definitive. Overall, this experiment serves both as an external validation of \textsc{DeepFact-Eval} and as further evidence that, as verifiers improve, residual disagreement on static factuality benchmarks increasingly reflects annotation divergence, ambiguity, or mis-scoped labels rather than true model errors. This in turn further motivates auditable, evolving evaluation protocols such as AtS.

\section{Conclusion}
We introduce DeepFact-Eval, an agentic verifier for deep research factuality, and DeepFact-Bench, an evolving benchmark for evaluating verifiers. We show that domain experts are unreliable as DRR labelers, motivating \emph{Audit-then-Score (AtS)}: an auditable, human–AI protocol where ground truth is a revisable consensus updated when challengers provide stronger evidence. Instantiating AtS with DeepFact-Bench, we show that DeepFact-Eval outperforms other fact-checkers. More broadly, evolving benchmarking offers a path to evaluation as AI approach or surpass expert-level performance.

\section*{Limitations}
Our current verifiers function as expert literature reviewers rather than active laboratory scientists, constrained to validating claims against \textit{existing} scientific literature. Consequently, they cannot empirically verify findings through new experiments or data simulations—a gap that future ``AI Scientists'' must bridge to address scenarios where the literature is silent or conflicted~\cite{lu2024ai}. Beyond these epistemic limits, significant opportunities for efficiency improvement remain. Although we introduce a lite variant of DeepFact-Eval, the necessity of long-context reasoning and iterative retrieval makes deep verification expensive for long-form reports. This computational burden currently limits real-time applicability, despite the framework's necessity for high-stakes accuracy. 

\section*{Ethic Considerations}
While our tools are designed to verify factuality, the underlying technology (generating and refining complex claims) could theoretically be repurposed to generate sophisticated factual inaccuracies. However, we believe the development of strong verification tools is the most effective countermeasure against such risks. By releasing DeepFact-Bench and DeepFact-Eval, we aim to empower the community to detect and refute hallucinated or manipulated scientific reports.

\bibliography{custom}

\appendix

\section{DRR Claims Factuality Definition}
\label{appendix:factuality_definition}
\subsection{DRR Claims Factuality Definition}
We adapt VeriScore’s atomic-fact definitions \cite{song-etal-2024-veriscore} of supported, contradictory, and inconclusive to the sentence level for DRR evaluation (\autoref{tab:drr_factuality_definitions}).
\begin{table*}[h]
\centering
\small
\setlength{\tabcolsep}{6pt}
\renewcommand{\arraystretch}{1.25}
\begin{tabular}{p{0.14\linewidth} p{0.48\linewidth} p{0.34\linewidth}}
\toprule
\textbf{Label} & \textbf{Definition (sentence-level)} & \textbf{Requirements / Decision criteria} \\
\midrule
\textbf{Supported} &
All factual claims in the sentence are directly or logically supported by at least one source, and no equally or more credible source contradicts any part of the sentence. &
(i) The source(s) cover every factual element of the sentence. \newline
(ii) If inference is used, it must be transparent and logically valid (no speculative leaps). \newline
(iii) Evidence is sufficient to guarantee the claim as stated. \newline
(iv) No equally or more credible source refutes any part of the sentence. \\

\textbf{Inconclusive} &
If no claims are \emph{Contradictory}, but at least one claim is \emph{Inconclusive}, and the rest are either \emph{Inconclusive} or \emph{Supported}, then the sentence is marked \emph{Inconclusive}. &
(i) At least one claim lacks direct support \emph{and} lacks credible refutation. \newline
(ii) The inference chain is weak/speculative or not clearly grounded. \newline
(iii) Evidence is missing or internally conflicting without a clear resolution. \\

\textbf{Contradictory} &
If any single factual claim in the sentence is contradicted by a reliable source, and no equally strong support exists, the entire sentence is marked \emph{Contradictory}. &
(i) At least one claim is clearly refuted (negated or directly contradicted) by a reliable source. \newline
(ii) No stronger or equally reliable evidence supports the refuted claim. \\

\textbf{None} &
The sentence contains no verifiable factual claims; it instead expresses opinions, vague speculation, moral judgments, or rhetorical language. &
Non-verifiable (see \autoref{tab:non_verifiable_categories}). \\
\bottomrule
\end{tabular}
\caption{DRR sentence-level factuality labels. A sentence aggregates over its constituent factual claims: any contradicted claim yields \emph{Contradictory}; otherwise, any unresolved claim yields \emph{Inconclusive}; otherwise \emph{Supported}; and \emph{None} if no verifiable factual claims are present.}
\label{tab:drr_factuality_definitions}
\end{table*}

\subsection{None-Verifiable Definition}
\label{appendix:non-verifiable}
We define what is non-verifiable types in DRR (\autoref{tab:non_verifiable_categories}) based on our in-house annotation findings.

\subsection {Deep Research Report Claim Error Taxonomy}
\label{appendix:error_taxonomy}
We observe several common error patterns of deep research models in our pilot annotations. And therefore we categorize errors as in \autoref{tab:error-taxonomy}

\section{Expert Annotations}
\label{appendix:expert_annotations}
\subsection{Pilot In-House Annotations}
\label{appendix:pilot_annotations}
To calibrate the difficulty of DRR claim verification, we first conducted a pilot round of in-house annotations. We prompted LLMs with research questions and collected reports generated by multiple deep-research agents, then attempted to verify the resulting claims against the global literature.

This pilot immediately revealed that DRR verification is qualitatively more demanding than standard claim checking. A single report can contain \emph{hundreds} of verifiable statements, and a single challenging claim can take \emph{hours} to resolve due to multi-hop dependency on scattered, technical sources. This makes exhaustive, claim-by-claim verification infeasible at scale.

\paragraph{Domain drift and fragmentation amplify burden and reduce reliability.}
We found that even slight domain drift sharply increases annotation time while degrading reliability. For example, asking a PhD-level annotator specializing in LLM RL to verify a report centered on RAG (or vice versa) often led to slower verification and more errors, despite both topics falling under the broad ``LLM'' umbrella. Similarly, expertise can decay with \emph{temporal drift}: an annotator who previously worked on RAG but has since shifted to agentic systems may be less familiar with the most recent literature, making verification substantially harder. These observations suggest that \emph{hyper-specialization} is a core obstacle for DRR verification: the effective competence set is narrow, and modest topic/time mismatches can push claims beyond a reasonable verification budget (e.g., hours per claim).

\paragraph{Multi-annotator adjudication is not a silver bullet.}
In this setting, conventional multi-annotator adjudication is often less informative than expected. When secondary annotators have even slight domain mismatch, they frequently defer to the primary annotator (or converge on the same surface-level judgment) due to limited confidence and familiarity, which can inflate agreement without improving correctness.

\paragraph{Cognitive load is extreme.}
Finally, we observed pronounced attention decay over long annotation sessions. Annotators can remain highly focused on the first several claims, but performance deteriorates as the report length and verification horizon grow---a predictable failure mode when the task involves hundreds of decisions with heavy context switching.

\paragraph{Design implications.}
These pilot findings directly shaped our full-scale annotation protocol. To reduce cognitive overload, we (i) developed an annotation interface with visualization and navigation support, and (ii) used importance- and risk-stratified sampling to concentrate effort on the most consequential and error-prone claims rather than attempting exhaustive verification. (iii) Since expert annotation is no longer reliable, we design micro-gold to quantify how reliable it is.

\begin{figure*}[t]
    \centering
    \includegraphics[width=0.95\linewidth]{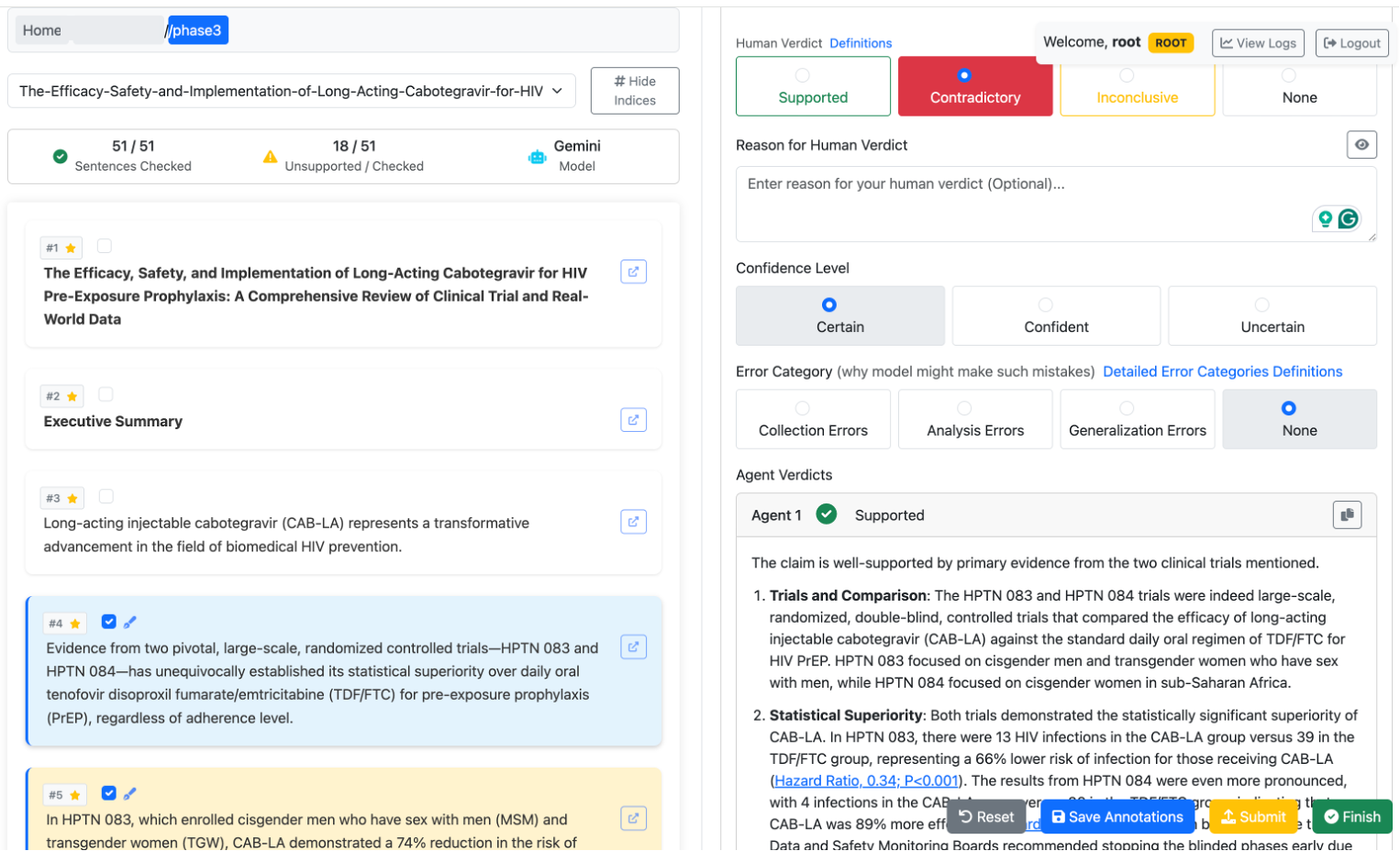}
    \caption{Annotation Interface for DRR}
    \label{fig:interface}
\end{figure*}

\begin{figure*}[h]
    \centering
    \begin{subfigure}[t]{0.45\textwidth}
        \centering
        \includegraphics[width=\linewidth]{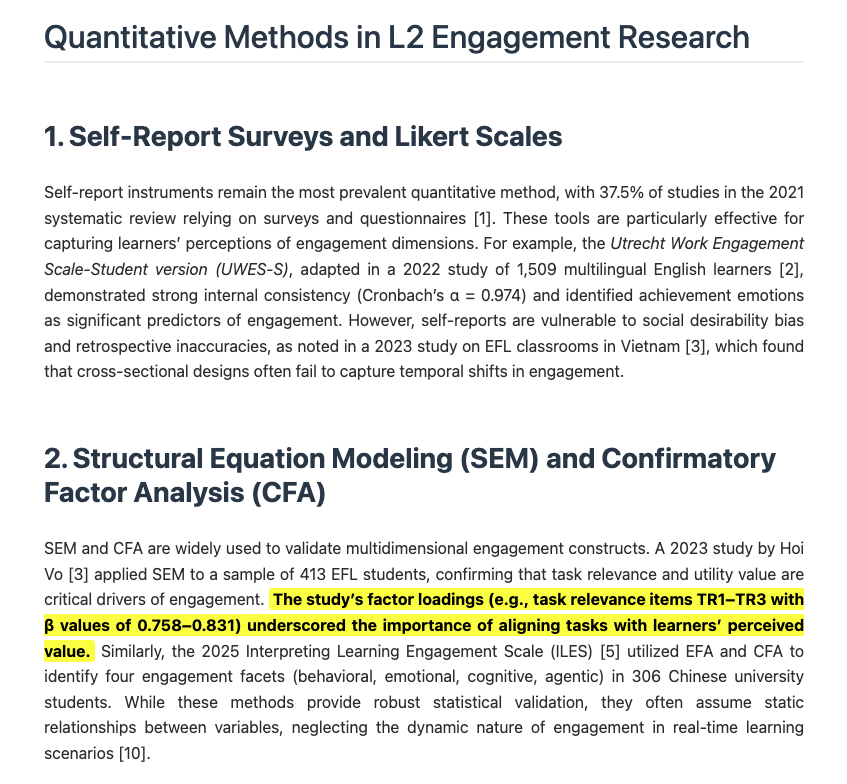}
        \caption{Jump from a claim to its exact span in the original report for fast, low-friction context recovery.}
        \label{fig:locate-claim}
    \end{subfigure}\hfill
    \begin{subfigure}[t]{0.45\textwidth}
        \centering
        \includegraphics[width=\linewidth]{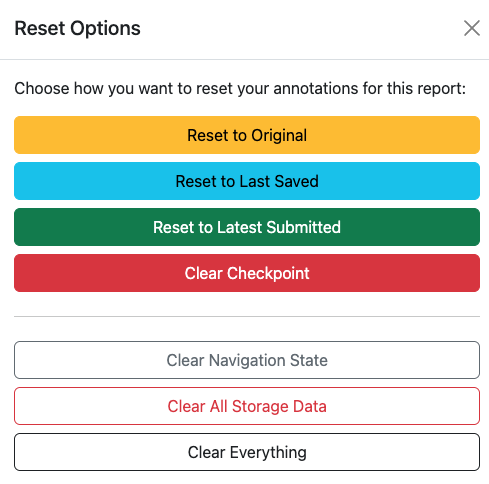}
        \caption{Reset to an earlier checkpoint to resume long-horizon annotation without losing progress.}
        \label{fig:reset-checkpoints}
    \end{subfigure}

    \caption{\textbf{Interface features.} Left: Jump from a claim to its exact span in the original report for fast, low-friction context recovery.
Right: Reset to an earlier checkpoint to resume long-horizon annotation without losing progress.}
    \label{fig:ui-features}
\end{figure*}
\subsection{Annotation Visualization}
To reduce cognitive overload, we built an annotation interface \autoref{fig:interface} that makes DRR reading and labeling lightweight.  Annotators first select from the set of reports assigned to them. For each report, the interface shows a clear progress indicator (e.g., how many claims remain). The report is segmented into sentences in the left panel, while the right panel contains the fields to complete: the human verdict and a brief justification. In phases where we provide model assistance, the UI also displays the agent(s)’ verdicts and rationales as reference. For claims marked Unsupported, annotators additionally choose an error category. Definitions of the factuality labels and error taxonomy are embedded in the interface for quick lookup.
Annotators can jump from any claim directly to its location in the original report for quick context recovery \autoref{fig:locate-claim}.
The UI also supports checkpointing: they can reset to earlier checkpoints and resume later, enabling long-horizon annotation without losing state and helping prevent fatigue-driven errors \autoref{fig:reset-checkpoints}.

\subsection{Expert Recruitment}
To mitigate domain drift, we recruit PhD-level experts through university channels and have them verify DRRs derived from \emph{their own research questions} within their specialization. We require annotators to be currently enrolled PhD students to ensure up-to-date familiarity with recent literature; in pilot audits, more senior researchers (e.g., faculty) were often less attuned to fine-grained details despite strong broad expertise. Experts must demonstrate domain competence (i.e, first-author publications and reviewer experience in the area) and complete a short general-domain calibration task to align on our factuality definitions (\autoref{tab:drr_factuality_definitions}). To discourage low-effort labeling and improve reliability, compensation is contingent on passing a hidden micro-gold quality check, incentivizing sustained attention throughout verification. Our experts span control theory, environmental engineering, education, public health, and engineering management.  In total, the annotation process required more than 400 expert-hours. Participants were told that their responses (verdicts and rationales) would be used to develop and evaluate DRR factuality benchmarks and may be released in de-identified form. We did not collect or release personally identifying information; all records were anonymized and stored under access control. 
Compensation averages about \$30/hour, and is region-adjusted and set above the local research-assistant hourly wage.

\subsection{Annotation Stage 1: DRR Generation}
\label{appendix:drr_generation}
This stage constructs challenging yet verifiable prompts that elicit deep research reports (DRRs) requiring multi-source reasoning, synthesis, and evidence integration---settings where agentic LLMs are more likely to fail. To ensure verifiability, each prompt must fall within the annotator's domain of expertise, so that a knowledgeable expert can assess correctness efficiently.

\paragraph{Prompt authoring.}
Each expert follows the steps below:
\begin{enumerate}
    \item \textbf{Choose niche subtopics.} Select 2--3 highly specific subtopics they have published on or know well (e.g., granular methods, datasets, or keywords from recent work).
    \item \textbf{Write research-style questions.} Draft 6 well-scoped prompts that are precise (not generic) and typically require synthesizing or comparing multiple sources rather than summarizing a single document.
    \item \textbf{Ensure factual verifiability.} Avoid (i) speculative/future-looking prompts (e.g., ``predict trends''), (ii) opinion-based prompts (e.g., ``is X good?''), and (iii) unrealistic or non-verifiable requests (e.g., proposing novel experiments without established evidence).
    \item \textbf{Design for grounded outputs.} Prompts should yield factual claims supported by public literature and verifiable via online resources (papers, datasets, official reports).
\end{enumerate}

\paragraph{LLM clarification.}
Given each question, we run GPT-4.1 to request clarifications and missing details, mimicking the ``question refinement'' step used by OpenAI deep-research. And we integrate the feedback to a more specific prompts. 

\paragraph{Report generation.}
For each expert's 6 questions, we generate DRRs using three deep-research systems: OpenAI DeepResearch (o3), Gemini Deep Research (Gemini-2.5-pro), and OpenDeepResearch (Qwen-32B). We generate two reports per system.

\paragraph{Report selection.}
Each expert selects three reports they feel most confident verifying---one from each system.

\subsection{Annotation Stage 2: DRR Annotation (Audit-then-Score)}
Experts label and justify sampled claims from their selected DRRs through multiple rounds of auditing:

\paragraph{Round 0 (Expert-only).}
Experts annotate claims independently to form an initial static benchmark. Meanwhile, we ask the expert to indicate their confidence level (among Certain, confident, uncertain).

\paragraph{Round 1 (Audit Agent 1: SmolAgent GPT-4.1).}
Experts review Agent 1's verdicts and rationales, adopting them when the agent provides stronger evidence or reasoning. For the current and all following process, they are not aware if the agents are good or not, they are informed that agent and their own annotations both have correct and incorrect, and they need to use their own judgment to decide

\paragraph{Round 2 (Audit Agent 2: DeepFactEval GPT-4.1).}
Experts audit a stronger verifier and update labels when its rationale is better supported or more coherent. From this round onward, experts also provide their own rationales when making updates.

\paragraph{Round 3 (Audit Agent 3: DeepFactEval GPT-5).}
Experts audit and incorporate revisions from the strongest verifier, producing the latest benchmark version.

For all stages, annotators are blinded to the agents’ quality and are explicitly informed that both agent predictions and human annotations can contain errors. They are instructed to use their own judgment when accepting, rejecting, or revising any verdict. Across rounds, experts repeatedly revisit the same claims, encouraging careful reconsideration rather than one-shot labeling.

\subsection{Annotation Stage 3: Post-hoc Quality Check}
\label{appendix:post_quality_check}
To assess the quality of the released benchmark (including the micro-gold set), we conduct a post-hoc quality check roughly one month after Stage 2 to reduce memorization effects.

\paragraph{Non-micro-gold items.}
Experts re-annotate with all agents' verdicts and rationales visible. We measure intra-expert consistency against their earlier decisions. \cite{zhao2025sciarena}.

\paragraph{Micro-gold items.}
Experts are shown the micro-gold items and their construction process, and indicate whether they agree with the micro-gold verdict; if not, they provide a rationale. This adds an additional validity check for the micro-gold set.

Overall, intra-expert consistency is 92.7\%, and the micro-gold confirmation rate is 99.3\%.

Following this procedure, we obtain a test set of 621 claims from 15 reports spanning five domains. Separately, our in-house annotation yields a validation set of 323 claims from 5 CS reports. Together, these form DeepFact-Bench.

\section{Implementations}
\label{appendx:implementation}

\subsection{Traditional Methods}
Methods such as VeriScore, SAFE, Fire, FactCheck-GPT operate at the atomic-claim level: they decompose each sentence into multiple atomic facts and output a verdict per fact, whereas our method outputs a single verdict per sentence. FIRE is also designed for atomic-claim verification but does not include a built-in decomposition step; therefore, we first extract atomic claims using a GPT-4.1 claim extractor and then run FIRE on these claims. To make all baselines comparable, we aggregate atomic-level verdicts into a sentence-level verdict.

We follow VeriScore’s three-way label space \{supported, inconclusive, contradictory\}. FIRE and FactCheck-GPT use \{true, not-enough-evidence, false\}, which we map to {supported, inconclusive, contradictory}, respectively. For VeriScore/FIRE/FactCheck-GPT, we aggregate atomic verdicts using the rule:
(i) if any atomic claim is contradictory, the sentence is contradictory;
(ii) else if any atomic claim is inconclusive, the sentence is inconclusive;
(iii) otherwise supported.
For final evaluation, we merge contradictory and inconclusive into unsupported to obtain a binary label space \{supported, unsupported\}. SAFE outputs \{supported, not supported\} plus an irrelevant label. We aggregate SAFE by marking a sentence as unsupported if any atomic claim is not supported, otherwise supported; if all atomic claims are labeled irrelevant, we count the sentence as an incorrect prediction.
To this end, we convert all methods to a sentence-level binary prediction, enabling fair comparison.

\subsection{Cost Estimation}
\label{appendix:cost_estimation}
We estimate cost for all deep-research methods using OpenAI API token prices. DeepFact-Eval uses GPT-4.1 for verification, but uses GPT-4.1 mini to summarize full documents to reduce cost; we convert GPT-4.1 mini token usage into GPT-4.1–equivalent cost by scaling by the corresponding price ratios. GPT-Researcher is evaluated under its default RAG setup, which retrieves relevant passages using an OpenAI embedding model before generation. Under OpenAI’s listed rates (as of Dec 23, 2025), GPT-4.1 costs \$2.00 / 1M input tokens and \$8.00 / 1M output tokens, and GPT-4.1 mini costs \$0.40 / 1M input tokens and \$1.60 / 1M output tokens.

\subsection{Hyper Parameters}
For \textsc{DeepFact-Eval}, we use the following inference hyperparameters: max steps=2 (maximum iterations), max queries=5 (queries per step), max sources=40 (maximum retrieved sources retained for synthesis), and max completion tokens=8192 per request.

\begin{table*}[t]
\small
\setlength{\tabcolsep}{2.5pt}
\begin{tabular}{p{2.5cm} p{0.8cm} p{3.5cm} p{0.8cm} p{3.0cm} p{0.6cm} p{3.7cm}}
\toprule
\textbf{Claim} & \textbf{Label} & \textbf{Evidence} & \textbf{Model} & \textbf{Model reason} & \textbf{New} & \textbf{Note} \\
\midrule
CHEK2 is not associated with breast cancer. &
T &
We genotyped these six tagSNPs in 1,577 postmenopausal breast cancer cases and 1,513 population controls, but found no convincing association between any common CHEK2 haplotype and breast cancer risk. The 1100delC mutation was rare in our Swedish population—0.7\% in cases and 0.4\% in controls—with a corresponding odds ratio for carriers versus noncarriers of 2.26 (95\% confidence interval, 0.99–5.15). &
F &
Clinical genetics literature treats \textit{CHEK2} as a moderate-penetrance breast cancer susceptibility gene (e.g., pathogenic variants like 1100delC increase risk); thus the absolute ``not associated'' claim is false. &
F &
\textbf{Label-evidence misalignment:} the evidence supports “common CHEK2 polymorphisms/haplotypes were not convincingly associated”, but it does not support “CHEK2 (as a gene) is not associated,” since a known rare variant shows increased risk in their data and prior work \\
\midrule
In young and middle-aged adults, current or remote uses of ADHD medications do not increase the risk of serious cardiovascular events. &
T &
RR for serious CV events: current vs nonuse $0.83$; new users $0.77$; conclusion: not associated with increased risk. &
F &
Argues newer long-horizon evidence suggests duration-dependent increases in broader CVD outcomes (e.g., hypertension/arterial disease; cardiomyopathy), so the categorical ``do not increase'' is contradicted. &
F &
\textbf{Newer evidence founded}:
Earlier work (e.g., Habel et al., JAMA 2011)  found no increase in serious acute cardiovascular events with ADHD medication use, but newer studies with longer follow-up and cumulative exposure analyses report a modest increase in cardiovascular disease risk.  e.g., Zhang et al., JAMA Psychiatry 2024) \\
\midrule
Bone marrow cells contribute to adult macrophage compartments. &
F &
After bone marrow transplantation, host macrophages expanded when donor development was compromised (suggesting limited donor contribution). &
T &
General immunology evidence supports bone-marrow-derived monocytes contributing to adult macrophage pools (tissue- and context-dependent), so the broad claim can be true. &
T &
\textbf{Overgeneralizing context-specific replenishment into a global macrophage origin claim: }Benchmark evidence supports \emph{minimal contribution} in a specific setting/tissue; claim is generic and may hold across tissues/conditions. \\
\bottomrule
\end{tabular}
\caption{\textbf{Disagreements of Annotation on SciFact.} We manually inspect instances where the model's prediction disagrees with the benchmark verdict. \textbf{T} = supported/true, \textbf{F} = unsupported/false (including contradictory/inconclusive), \textbf{U} = unverifiable. ``\texttt{label}'', ``\texttt{model}'', and ``\texttt{new}'' correspond to the original dataset label, the model’s predicted label, and the expert re-annotated label, respectively. ``\texttt{evidence}'', ``\texttt{model\_reason}'', and ``\texttt{note}'' correspond to the SciFact-provided rationale (abstract sentences used to justify the original label), the model's rationale, and the expert re-annotation rationale, respectively; all are summarized concisely.}
\label{tab:scifact-disagreements}
\end{table*}

\section{Results on Other Datasets}
\label{appendix:other_datasets}
We evaluate \textsc{DeepFact-Eval} beyond our benchmark on three established factuality datasets: SciFact~\cite{wadden-etal-2020-scifact}, ExpertQA~\cite{malaviya-etal-2024-expertqa}, and Factcheck-Bench~\cite{wang-etal-2024-factcheck-gpt}.

\subsection{Dataset Set-up}
\paragraph{SciFact.}
SciFact~\cite{wadden-etal-2020-scifact} is a scientific claim verification benchmark where claims are derived from citation sentences (\emph{citances}) and verified against a corpus of paper abstracts. Each $(\text{claim}, \text{abstract})$ pair is labeled \textsc{Supports}, \textsc{Refutes}, or \textsc{Noinfo}, and \textsc{Supports}/\textsc{Refutes} instances include gold rationale sentences from the abstract. We evaluate on the SciFact validation set under a binary setting by excluding \textsc{Noinfo} and treating the task as \textsc{Supports} vs.\ \textsc{Refutes}/\emph{unsupported}, yielding $188$ evaluated instances; \textsc{DeepFact-Eval} disagrees with the gold label on $29/188$ (15.4\%).

\paragraph{ExpertQA.}
ExpertQA~\cite{malaviya-etal-2024-expertqa} pairs expert-authored questions with LLM responses. Responses are split into sentences, and annotators assign a sentence-level factuality label on a five-point scale (\emph{Definitely correct / Probably correct / Unsure / Likely incorrect / Definitely incorrect})~\cite{malaviya-etal-2024-expertqa}. We focus on domains where we can access relevant experts---Engineering \& Technology, Education, Environmental Science, and Healthcare/Medicine---and restrict to the most objective labels (\emph{Definitely correct} and \emph{Definitely incorrect}). We sample 100 per label (200 total); \textsc{DeepFact-Eval} disagrees with ExpertQA on $90/200$ (45\%).

\paragraph{Factcheck-Bench.}
Factcheck-Bench is built from LLM-generated answers to open-domain questions (in-house questions, Dolly Closed-QA, Dolly Open-QA). The authors decompose these responses into 678 claims, label 661 as checkworthy, and include 94 $(\text{question}, \text{response})$ pairs~\cite{wang-etal-2024-factcheck-gpt}. Following FIRE's cross-dataset preprocessing~\cite{xie-etal-2025-fire}, we map the original four labels (\emph{supported, partially supported, not supported, refuted}) to a binary scheme: supported/partially supported $\rightarrow$ \textsc{True}, refuted $\rightarrow$ \textsc{False}, and exclude \emph{not supported}. This yields 631 labeled claims (472 \textsc{True}, 159 \textsc{False}). We sample 200 claims for evaluation; \textsc{DeepFact-Eval} disagrees with the benchmark on $36/200$ (18\%) cases.

\subsection{Expert Re-annotation Methods}
We audit only the \emph{disagreements} between \textsc{DeepFact-Eval} and each benchmark and assign each case to a subject-matter expert in the closest matching domain.

\paragraph{SciFact: evidence-grounded two-stage audit.}
Because SciFact includes abstracts and gold rationale sentences, we audit each disagreement in two stages:
\begin{enumerate}
    \item \emph{Evidence--label consistency check:} verify whether the SciFact-provided rationale/evidence (abstract sentences) actually entails the gold verdict.
    \item \emph{Blind rationale adjudication:} for the remaining cases, present experts with two blinded packages---(i) SciFact rationale + abstract and (ii) \textsc{DeepFact-Eval} rationale + retrieved abstract---and ask which explanation is better supported and why.
\end{enumerate}

\paragraph{ExpertQA and Factcheck-Bench: blinded re-annotation on a subset.}
ExpertQA and Factcheck-Bench do not provide per-claim, evidence-grounded rationales, which makes disagreement adjudication harder. We therefore rely on blinded re-annotation for disputed claims (experts in the closest domains for ExpertQA; authors for Factcheck-Bench). For ExpertQA, we do not re-annotate all disagreements: we first screen all disputed items to flag clear labelability issues (e.g., non-verifiable discourse sentences). Among the remaining 76 disagreements, we randomly sample 30 claims for blinded re-annotation, where annotators do not see either the dataset label or the model prediction.

\subsection{Results}
\paragraph{SciFact.}
Among the $29$ disagreements: (i) $12/29$ arise from evidence–label misalignment, where the provided abstract does not substantiate the annotated verdict; (ii) among the remaining 17, blind adjudication finds $7/17$ are unresolvable due to insufficient expert confidence; and (iii) of the $10$ resolvable cases, experts favor the model’s interpretation in $4/10$. Extrapolating this preference rate to all 17 adjudicated cases and evaluating against the resulting expert-recalibrated labels, \textsc{DeepFact-Eval} achieves an estimated $178/188$ accuracy (94.7\%), suggesting that a substantial fraction of the apparent errors may be driven by annotation noise rather than systematic model failures. (Examples in \autoref{tab:scifact-disagreements})

\paragraph{ExpertQA.}
Out of $90/200$ disagreements, the author inspection suggests that $14/90$ are non-verifiable discourse sentences (e.g., hedges, conversational prompts, generic advice) that arguably should not receive factuality labels. For the remaining disagreements, we sample 30 claims for blind expert re-annotation, allowing the experts to use search engines and LLM tools; experts side with \textsc{DeepFact-Eval} in $28/30$ cases and with the original dataset label in $2/30$ (see examples in \autoref{tab:expertqa-factcheckbench-disagreements}). However, because ExpertQA does not provide per-claim rationales, many conflicts are difficult to adjudicate conclusively.

\paragraph{Factcheck-Bench.}
\textsc{DeepFact-Eval} disagrees with the benchmark on $36/200$ (18\%) cases. Manual inspection suggests $32/36$ disagreements likely reflect annotation noise (e.g., subjective/ambiguous claims, or cases where the author's founded evidence aligns more with the model judgment than the benchmark label; examples in \autoref{tab:expertqa-factcheckbench-disagreements}). Similarly, since the dataset does not provide rationales, these disagreements are difficult to resolve conclusively.

\paragraph{Summary.}

We summarize these findings using a three-way taxonomy to distinguish the source of the conflict:
\begin{enumerate}
    \item \textbf{Agreement}: the verifier matches the benchmark label.
    \item \textbf{Disagreement: Annotation divergence}: the verifier does not match the benchmark label, and our re-annotation diverges from the benchmark label as well, which indicates the disagreement cannot be cleanly resolved into a definitive model error. This includes evidence--label misalignment (SciFact), non-verifiable or non-checkworthy sentences being labeled (ExpertQA), subjective or underspecified claims, and cases that cannot be conclusively adjudicated due to missing gold rationales (ExpertQA, Factcheck-Bench).
    \item \textbf{Disagreement: Likely model error}: the verifier does not match the benchmark label, and our re-annotation aligns with the benchmark label, suggesting a likely verifier error.
\end{enumerate}

For ExpertQA and SciFact, the proportions for \emph{Annotation Divergence} and \emph{Model Error} are estimated by extrapolating the ratios observed in our annotated subsets to the total number of disagreements.

Results in \autoref{fig:other_datasets} reveals a consistent performance saturation point: as verifiers mature, residual discrepancies are driven predominantly by labeling noise and claim ambiguity rather than model error. These edge cases are difficult to adjudicate without evidence-grounded rationales. This underscores the necessity for auditable, evolving benchmarking, which allows for the diagnosis and correction of data artifacts, disentangling them from genuine verification failures.

\begin{table*}[htbp]
\small
\setlength{\tabcolsep}{3pt}
\begin{tabular}{p{3.2cm} p{1.0cm} p{1.0cm} p{4.8cm} p{0.6cm} p{4.0cm}}
\toprule
\textbf{Claim} & \textbf{Label} & \textbf{Model} & \textbf{Model reason (concise)} & \textbf{New} & \textbf{Note (concise)} \\
\midrule

\multicolumn{6}{l}{\textbf{ExpertQA}} \\
\midrule
Perhaps you could narrow down your question by specifying a time period, a location, or a source of information that you are interested in. &
F &
T &
Treats the suggestion as “supported” by research-methods best practices (narrowing questions by time/location/source). &
U &
This is conversational advice, not a checkable world fact. \\
\midrule
Website security: You should verify that your website is safe and trustworthy for your users. &
F &
T &
Best-practice guidance (e.g., OWASP/NIST/CISA-style) recommends validating security to protect users and maintain trust. &
T &
Normative, but still a valid best-practice claim; “supported” fits if ExpertQA treats recommendations as verifiable. \\
\midrule
An in-memory database is a database that resides entirely in the memory of the application and does not persist data on disk. &
T &
F &
In-memory DBs store primarily in RAM, but many support disk persistence (snapshots/logging) and are not necessarily “in the app’s memory.” &
F &
Overly absolute / incorrect definition due to “entirely” and “does not persist.” \\

\midrule
\multicolumn{6}{l}{\textbf{FactCheck-Bench}} \\
\midrule
Abacus computing doesn't need large amounts of energy. &
F &
T &
An abacus is manual/mechanical and requires no electricity; energy use is only minimal human effort, negligible vs. electronic computing. &
T &
True in the intended “no external power / low energy” sense. \\
\midrule
The decision on \emph{NYSRPA v. Bruen} was seen as a setback for gun rights supporters. &
T &
F &
\emph{Bruen} struck down New York’s “proper cause” carry-permit requirement and was widely framed as a win for gun rights advocates; calling it a setback for them is incorrect. &
F &
Reverses the typical framing (setback for gun control efforts, not for gun rights supporters). \\
\midrule
China's Olympic success has been remarkable since 1980s. &
F &
T &
Since first full participation in 1984, China has sustained high medal counts and frequent top medal-table finishes, supporting “strong success since the 1980s.” &
U &
“Remarkable” is subjective; better treated as non-verifiable despite strong supporting facts. \\
\midrule
Lawson was first established in 1975. &
T &
F &
Ambiguous entity: \emph{Lawson, Inc. (Japan)} dates to 1975, but the \emph{Lawson brand/origin} traces to a U.S. dairy store (1939). Without “in Japan,” the absolute claim is misleading. &
F &
Underspecified; correct only under a narrower reading (“Lawson, Inc. in Japan”). \\
\bottomrule
\end{tabular}
\caption{\textbf{Disagreements of Annotation on ExpertQA and FactCheck-Bench.} We manually reannotate and inspect instances where the model's prediction disagrees with the benchmark verdict. \textbf{T} = supported/true, \textbf{F} = unsupported/false (including contradictory/inconclusive), \textbf{U} = unverifiable. ``\texttt{label}'', ``\texttt{model}'', and ``\texttt{new}'' correspond to the original dataset label, the model’s predicted label, and the expert re-annotated label, respectively.}
\label{tab:expertqa-factcheckbench-disagreements}
\end{table*}

\section{Importance- and Risk-Stratified Claim Sampling}
\label{appendix:imp_risk_sampling}

Deep research reports frequently contain hundreds or thousands of distinct
claims, making exhaustive annotation infeasible.  We therefore design a
\textbf{two-factor sampling scheme} that emphasises
(i)~\emph{importance}—how central a claim is to the report’s thesis(defined in \autoref{tab:claim_importance_scale})—and
(ii)~\emph{risk}—the probability that the claim is incorrect according to an
automatic evaluator (SmolAgent with GPT-4.1).  By concentrating annotation effort on the most
consequential and most error-prone statements, we obtain a balanced yet
information-dense subset of claims.

\begin{table*}[h]
\centering
\small
\begin{tabular}{p{3cm}p{11cm}}
\toprule
\textbf{Score} & \textbf{Definition} \\
\midrule
\textbf{5 -- Backbone claim} & Essential to the core thesis. Removing this would break the logic or invalidate the main takeaway. Often appears in the title, abstract, or conclusion, and is referenced multiple times. \\
\textbf{4 -- Critical support} & Key evidence or reasoning that directly supports a backbone claim. Removing it weakens the argument substantially but does not break the core conclusion. \\
\textbf{3 -- Standard support} & Provides helpful background or secondary evidence. Removing it moderately weakens the report but leaves the main thesis intact. \\
\textbf{2 -- Minor context} & Background detail, definition, or peripheral comment. Removing it has little to no effect on the main message. \\
\textbf{1 -- Irrelevant or off-topic} & Unrelated, redundant, or likely a segmentation artifact. Removing it improves clarity or focus. \\
\bottomrule
\end{tabular}
\caption{Claim importance scale used for prioritizing verification.}
\label{tab:claim_importance_scale}
\end{table*}

\paragraph{Step 1 – Quota definition.}
Given a target batch size \(N\), we allocate per-bucket quotas over five
importance levels, e.g.\ \(\{5{:}\,40\%, 4{:}\,35\%, 3{:}\,20\%,
2{:}\,5\%, 1{:}\,0\%\}\).
For level \(i\), the quota is \(q_i=\lfloor N \times p_i\rfloor\) where
\(p_i\) is the desired proportion.
The quotas satisfy \(\sum_i q_i=N\).

\paragraph{Step 2 – Risk weights.}
Each candidate claim is tagged by an automatic factuality evaluator as
\textsc{Supported} or \textsc{Unsupported}/\textsc{Low-Confidence}.
We assign a risk weight
\[
  w_j \;=\;
  \begin{cases}
     1,      & \textsc{Supported}\\[2pt]
     \rho>1, & \textsc{Unsupported}
  \end{cases}
\]
where \(\rho\) controls how strongly
we oversample likely errors.

\paragraph{Step 3 – Quota adjustment for sparse buckets.}
If an importance bucket contains fewer than \(q_i\) candidates, we
down-scale \(q_i\) to the available count and redistribute the deficit
proportionally across buckets that still have surplus capacity.  This
guarantees the final sample size remains exactly \(N\).

\paragraph{Step 4 – Risk-weighted sampling without replacement.}
Within each bucket we sample without replacement, using inclusion
probability
\[
  \Pr(j \mid j\in i)=
  \frac{w_j}{\sum_{k\in i} w_k}.
\]

\begin{table*}[h]
\centering
\small
\setlength{\tabcolsep}{6pt}
\renewcommand{\arraystretch}{1.2}
\begin{tabular}{p{0.05\linewidth} p{0.20\linewidth} p{0.20\linewidth} p{0.25\linewidth} p{0.20\linewidth}}
\toprule
\textbf{\#} & \textbf{Category} & \textbf{Typical Location} & \textbf{Why It's Non-Verifiable} & \textbf{Concrete Examples} \\
\midrule
A & Document-structure \& rhetorical framing (headings, previews, transitions)
& Section titles, figure headings, opening or bridging sentences
& Merely label or orient the reader; they don't assert external facts.
& ``3 Dataset Construction''; ``In the next subsection we detail the ablation study.'' \\
\addlinespace

B & Forward-looking statements
& Abstract, introduction, future-work, conclusion
& Refer to intentions or events that have not happened yet.
& ``We will extend this method to multilingual data next year.'' \\
\addlinespace

C & Research questions \& hypotheses
& Introduction, methods
& Pose uncertainties; they explicitly seek answers rather than state facts.
& ``Does larger model size improve calibration?'' \\
\addlinespace

D & Subjective judgments \& opinions
& Discussion, conclusion, related-work critiques
& Depend on author viewpoint or value judgments.
& ``Our approach is considerably more elegant than prior work.'' \\
\addlinespace

E & Citation lists \& bibliographic metadata
& In-text citations, References section
& Point to external sources; correctness is about formatting, not truth value.
& ``(Smith et al., 2024)'' \\
\addlinespace

F & Speculative or motivational claims
& Introduction, broader-impact, abstract
& Describe possibilities, potential impact, or high-level vision rather than established facts.
& ``This technique could revolutionize personalized medicine.'' \\
\bottomrule
\end{tabular}
\caption{Common types of non-verifiable sentences in DRRs (mapped to the \texttt{None} label).}
\label{tab:non_verifiable_categories}
\end{table*}

\section{Audit-then-Score Algorithm}
\label{appendix:ats}
The full algorithm of AtS is listed below in Algorithm \ref{alg:ats}
\begin{algorithm*}[htbp]
\caption{The Audit-then-Score (AtS) Protocol}
\label{alg:ats}
\begin{algorithmic}[1]
\Statex \textbf{Prerequisite:} An initial seed benchmark $B_0 = \{(c_i, d_i, y_i^{(0)}, \rho_i^{(0)})\}_{i=1}^N$ created by human experts.
\Statex
\Procedure{EvolveBenchmark}{$B_t, M_t, A_t$}
    \Comment{$B_t$: current benchmark, $M_t$: Challenger, $A_t$: Auditor}
    \State $U_{M_t,t} \gets \emptyset$ \Comment{Initialize empty proposal}
    \State $\hat{Y} \gets \emptyset$ \Comment{Initialize set of all model predictions}

    \Statex \Comment{\textbf{Phase 1: Generate Verdicts and Challenges}}
    \For{each claim $i$ from $1$ to $N$}
        \State $(c_i, d_i, y_i^{(t)}, \rho_i^{(t)}) \gets B_t[i]$ \Comment{Get current benchmark data}
        \State $(\hat{y}_i, \hat{\rho}_i) \gets M_t(c_i, d_i)$ \Comment{Run Challenger model}
        \State $\hat{Y} \gets \hat{Y} \cup \{(i, \hat{y}_i)\}$ \Comment{Store all predictions for final scoring}
        
        \If{$\hat{y}_i \neq y_i^{(t)}$}
            \State $U_{M_t,t} \gets U_{M_t,t} \cup \{(i, \hat{y}_i, \hat{\rho}_i, \rho_i^{(t)})\}$ \Comment{Add disagreement to proposal}
        \EndIf
    \EndFor

    \Statex \Comment{\textbf{Phase 2: Audit}}
    \State $\Delta B_t \gets \emptyset$ \Comment{Initialize empty set of updates}
    \For{each challenge $(i, \hat{y}_i, \hat{\rho}_i, \rho_i^{(t)})$ in $U_{M_t,t}$}
        \If{$A_t(\hat{\rho}_i, \rho_i^{(t)}) = \texttt{ACCEPT}$} \Comment{Auditor adjudicates}
            \State $\Delta B_t \gets \Delta B_t \cup \{(i, \hat{y}_i, \hat{\rho}_i)\}$ \Comment{Accept the challenger's update}
        \EndIf
    \EndFor

    \Statex \Comment{\textbf{Phase 3: Evolve Benchmark}}
    \State $B_{t+1} \gets B_t \oplus \Delta B_t$ \Comment{Apply updates to create the new benchmark version}

    \Statex \Comment{\textbf{Phase 4: Score}}
    \State $Y^{(t+1)} \gets \{(i, y_i^{(t+1)}) \mid (c_i, d_i, y_i^{(t+1)}, \rho_i^{(t+1)}) \in B_{t+1}\}$ \Comment{Get new ground truth labels}
    \State $S \gets \Call{CalculateScore}{\hat{Y}, Y^{(t+1)}}$ \Comment{e.g., Accuracy}
    
    \State \Return $B_{t+1}, S$
\EndProcedure
\end{algorithmic}
\end{algorithm*}

\section{Statistical Significance of Findings}
\label{sec:significance}

As the DeepFact-Bench test set consists of 15 reports from multiple domains, which may introduce substantial cross-report variance, we additionally assess whether our main conclusions are robust to \emph{report sampling}. In particular, we test the significance of two central findings: (i) \textsc{AtS} improves human annotation quality, and (ii) \textsc{DeepFact-Eval} outperform existing verifier baselines.

\paragraph{Report-level paired bootstrap.}
We treat each \emph{report} as the independent unit, since claims within the same report are correlated and should not be treated as independent samples. To quantify uncertainty, we use a \emph{paired cluster bootstrap} over reports. For each comparison between two methods $A$ and $B$ (e.g., Round-3 vs.\ Round-2 human labels, or \textsc{DeepFact-Eval} vs.\ GPT-Researcher), we perform 20{,}000 bootstrap replicates. In each replicate, we sample 15 test reports \emph{with replacement}, recompute the \emph{micro-accuracy} of both methods on the same resampled report set, and record the paired difference
\[
d = \mathrm{score}(A) - \mathrm{score}(B).
\]
We then compute a 95\% confidence interval from the empirical bootstrap distribution of $d$. If the 95\% confidence interval excludes $0$, we consider the improvement statistically significant at approximately the 0.05 level.

\paragraph{AtS significantly improves human annotation quality across rounds.}
We first apply this procedure to human annotation accuracy on the micro-gold set across \textsc{AtS} rounds. The results show that human annotation quality improves significantly over time. For example, \textbf{Round-3} outperforms \textbf{Round-2} by \textbf{4.9 points}, with a \textbf{95\% confidence interval of [1.4, 7.9]}, which excludes $0$, confirming that the improvement in human label quality under \textsc{AtS} is not driven by a small subset of reports.

\paragraph{DeepFact-Eval significantly outperforms existing verifiers.}
We next compare \textsc{DeepFact-Eval} against existing verifier baselines using the same paired report-level bootstrap. \textsc{DeepFact-Eval} outperforms \textbf{GPT-Researcher} by \textbf{14.7 points} (\textbf{95\% CI: [7.4, 23.3]}) and \textbf{Smolagents} by \textbf{15.0 points} (\textbf{95\% CI: [9.5, 20.5]}). These results indicate that our main verifier gains are statistically robust and not driven by idiosyncrasies of a few sampled reports.

\paragraph{Takeaway.}
Overall, these report-level significance tests strengthen our conclusions in two ways. First, they show that \textsc{AtS} yields genuine improvements in human annotation quality. Second, they show that \textsc{DeepFact-Eval} and significantly outperform existing verifiers. Together, these results suggest that our findings are stable despite the limited number of reports in the current benchmark.

\section{Managing Evolving Benchmarks with AtS}
\label{appendix:manage_benchmark}
An evolving benchmark also requires explicit maintenance policies for governance, stopping, and fair reporting over time. In our setting, benchmark maintainers are responsible for curating updates, releasing new versions, and publishing changelogs of accepted revisions, making benchmark evolution transparent and auditable. 
To reduce drift toward verifier or agent auditor biases, we adopt two safeguards: hidden micro-gold monitoring to detect degradation, and periodic human recalibration once agent-driven updates exceed a small threshold (e.g., \(\sim\)5\% of benchmark verdicts). 
AtS is not intended to evolve indefinitely. In practice, stopping can be determined by one or more criteria: (i) a fixed audit budget, (ii) stabilization of micro-gold accuracy above a target threshold, and/or (iii) a preset maintenance horizon. Because AtS produces benchmark versions \(B_t\), fair comparison also requires versioned reporting: results should always specify the benchmark version, and longitudinal comparisons should be made either on frozen snapshots or by re-scoring archived outputs under an explicitly specified \(B_t\).

\section{More Related Work}
\label{appendix:more_related_work}
\paragraph{Dynamic Benchmarking.}
Dynamic benchmarking is “dynamic” in what changes over time: (i) the test set is iteratively expanded to track model weaknesses via human/model-in-the-loop adversarial rounds—Dynabench frames benchmarking as continuous data creation \cite{kiela-etal-2021-dynabench}, ANLI operationalizes this with iterative adversarial collection so the target keeps moving as models improve \cite{nie-etal-2020-adversarial}, real-time factuality assessment adversarially modify claims from news to make it difficult to fact-check \cite{chen-etal-2025-real} (ii) the world state changes, so benchmarks refresh questions/answers to stay current—FreshQA explicitly targets fast-changing knowledge and commits to regular updates \cite{vu-etal-2024-freshllms}, while RealTime QA evaluates on newly announced (e.g., weekly) questions tied to recent events \cite{kasai2023realtime}.  (iii) the evaluation distribution is mined “in the wild” and re-versioned, as in FactBench \cite{bayat2025factbench}, which curates prompts from real user interactions and is designed to be regularly updated with newly observed hallucination-triggering prompts.  (iv) the benchmark is refreshable by construction, where a repeatable generator yields new tasks—LiveDRBench \cite{java2025LiveDRBench} proposes “problem inversion” as a recipe to periodically produce new deep-research queries from existing reasoning problems.  Compared to these “refresh the inputs” approaches, our evolving benchmarking is dynamic primarily in the supervision itself: the benchmark’s ground truth and coverage are iteratively strengthened through auditing/verification (with ongoing quality control), not merely by swapping in new questions or sampling a new prompt stream

\paragraph{Expert-Led Benchmarking for Research Tasks.} Benchmarks for research tasks evaluate how LLM agents search, read, and synthesize literature, spanning tasks from literature-review \cite{openscholar} to Expert-level QA \cite{malaviya-etal-2024-expertqa, zhao2025sciarena}. Most existing evaluations implicitly treat expert(s) judgments as an infallible gold standard \cite{sharma2025researchrubrics, wang2025liveresearchbench, ruan2025expertlongbench}. Yet expert reliability is rarely quantified directly; instead, it is usually approximated via inter-annotator agreement, which obscures unresolved disagreements \cite{malaviya-etal-2024-expertqa, zhao2025sciarena} and cannot detect shared blind spots \cite{sharma2025researchrubrics, wang2025liveresearchbench}. Moreover, some benchmarks rely on STEM practitioners as annotators \cite{malaviya-etal-2024-expertqa, sharma2025researchrubrics} rather than the hyper-specialized researchers the questions may demand---further weakening the “ground truth.” This expert-dominance assumption will become a bottleneck as agents approach expert-level performance: the ceiling is set by annotation quality, and models may be penalized for correct outputs that conflict with noisy labels. Indeed, prior work reports that experts are not error-free, including annotation mistakes in HLE \cite{hle} and documented failures in human literature review \cite{salvador-olivan2019errors}. The issue is even more acute for verifying DRR, where a well-informed judgment may require locating and integrating substantial portions of the relevant literature. Our work challenges the assumption of human dominance by using adversarial hidden sets to monitor expert quality and proposing a human-AI collaborative framework to elevate benchmarking beyond expert limits. 

\paragraph{Role-based and multi-agent LLM systems.}
Role-based and multi-agent LLM systems are increasingly used as \emph{test-time scaffolds} for better task solving. Prior work assigns agents different roles \cite{qian-etal-2024-chatdev} or interaction protocols---e.g., role-playing cooperation in \textsc{CAMEL} \cite{li2023camel}, programmable multi-agent conversations in \textsc{AutoGen} \cite{wu2024autogen}, SOP-style collaborative workflows in \textsc{MetaGPT} \cite{hong2024metagpt}, and debate-based reasoning with multiple model instances \cite{icml-multiagent-debate, icml-multiagent-debate-factuality, chen-etal-2024-reconcile}. However, in these settings, the multi-agent system remains part of the \emph{solver}: it is designed to generate a better answer against a \emph{fixed, human-defined target}, such as a reference answer, solution, or rubric.
In contrast, our \textsc{AtS} framework is not merely a solver-side method; it is part of the \emph{benchmark}. The evaluated deep-research agent contributes the candidate claims/evidence, and the benchmark’s labels are updated through human--AI auditing rather than assumed static. This reframes the problem from ``using multiple agents to solve a task'' to ``using human--AI collaboration to keep evaluation trustworthy'' when systems approach or surpass expert-level capability.

\paragraph{Fact-Checking.} Check survey papers \cite{guo-etal-2022-factcheck-survey, hardalov-etal-2022-misinfo-survey} for more relevant work related to fact-checking.

\begin{table*}[h]
\centering
\small
\begin{tabular}{p{1.5cm} p{2.5cm} p{3.5cm} p{7.5cm}}
\toprule
\textbf{Code} & \textbf{Principle} & \textbf{Name} & \textbf{Definition / Example} \\
\midrule
\multicolumn{4}{c}{\textbf{1. Collection-Stage Errors (Evidence Gathering)}} \\
\midrule
C-AU & Authenticity & Fabricated Source & Cites a source, author, or quote that does not exist. \newline \textit{e.g., “OpenAI’s GPT-4V did ...” (no such study)} \\
C-PV & Provenance & Mis-sourced Evidence & Real fact but assigned to the wrong author, venue, or year. \newline \textit{e.g., arXiv preprint claimed to be a 2023 Nature paper} \\
C-CP & Completeness & Omitted Counter-Evidence & Omits accessible contradictory or qualifying evidence. \newline \textit{e.g., Ignores a larger meta-analysis contradicting a cited RCT} \\
C-CU & Currency & Out-of-Date Source & Relies on retracted or outdated sources without caveats. \newline \textit{e.g., Citing a 2019 draft despite a reversed 2024 version} \\
C-RE & Representativeness & Biased Sampling & Uses narrow evidence (e.g., language, geography) that skews conclusions. \newline \textit{e.g., All English news used to infer global media trends} \\
C-CX & Contextual Relevance & Contextual Mismatch & Collects evidence topically related but from a different domain or task. \newline \textit{e.g., Legal claim supported using biomedical QA accuracy} \\
\midrule
\multicolumn{4}{c}{\textbf{2. Analysis-Stage Errors (Evidence Processing)}} \\
\midrule
A-N1 & Numerical Fidelity & Numeric Distortion & Misrepresents counts, percentages, means, or CIs. \newline \textit{e.g., 25\% vs. 0.25 absolute points} \\
A-S1 & Semantic Fidelity & Semantic/Entity Swap & Substitutes similar but non-equivalent terms (e.g., metric, dataset type, model variant). \newline \textit{e.g., “faithfulness” reported when only F1 was measured} \\
A-P1 & Causal Discipline & Causal Projection & Claims causality from correlation or reverses direction. \newline \textit{e.g., “Retrieval reduces hallucination” based on observational data} \\
A-X1 & Study Integrity & Cross-Study Conflation & Blends results from different studies into a single narrative. \newline \textit{e.g., Claims KnowPO outperforms CTPC with no direct comparison} \\
A-B1 & Balanced Synthesis & Cherry-Picked Synthesis & Selects supportive evidence while omitting stronger contradictory data. \newline \textit{e.g., Cites positive RCT, ignores null meta-analysis} \\
A-T1 & Temporal Alignment & Temporal Misalignment & Compares studies/data from incompatible timeframes. \newline \textit{e.g., Comparing 2018 vs. 2024 SQuAD results} \\
A-O1 & Aggregation Soundness & Over-Aggregation & Combines incompatible metrics or tasks into a single number. \newline \textit{e.g., Merging latency, accuracy, and cost into one score} \\
A-C1 & Logical Coherence & Contradiction Ignorance & Presents contradictory findings without resolving them. \newline \textit{e.g., Quotes studies with opposite trends as co-validating} \\
A-L1 & Reasoning Validity & Chain-of-Thought Leap & Introduces an unjustified intermediate premise. \newline \textit{e.g., “Since large models are always calibrated...” (unsupported)} \\
\midrule
\multicolumn{4}{c}{\textbf{3. Generalization-Stage Errors (Claim Expansion)}} \\
\midrule
G-O1 & Scope Discipline & Over-Scope Leap & Generalizes beyond the evidence’s domain, task, or population. \newline \textit{e.g., From WebQA to biomedical QA without evidence} \\
G-H1 & Claim Proportionality & Hyperbolic Statement & Turns conditional or limited findings into absolutes. \newline \textit{e.g., “Always improves performance”} \\
G-T1 & Taxonomic Completeness & Taxonomy Oversimplification & Omits known categories or claims exhaustiveness without support. \newline \textit{e.g., “Two types of evaluation” ignoring a third} \\
G-C1 & Condition Transparency & Conditional Collapse & Drops necessary qualifiers or assumptions. \newline \textit{e.g., Removes “in low-resource settings” from claim} \\
G-R1 & Temporal Projection & Recency Extrapolation & Projects recent trend into the future without evidence. \newline \textit{e.g., 3-month rise to “will keep increasing exponentially”} \\
G-B1 & Base-Rate Awareness & Base-Rate Neglect & Reports large relative gains on near-zero baselines. \newline \textit{e.g., “50\% gain in recall” where base rate is 0.2\%} \\
G-S1 & Evidentiary Sufficiency & Single-Study Certainty & Claims general truth from one small study. \newline \textit{e.g., Lab study to industry-wide claim} \\
\bottomrule
\end{tabular}
\caption{Taxonomy of factuality errors in deep research report generation, organized by cognitive phase. Each code reflects a distinct violated principle.}
\label{tab:error-taxonomy}
\end{table*}

\section{Use of Ai Assistants}
We used LLMs to assist with writing. Specifically, we employed GPT-5 thinking, GPT-5 and GPT-4o to rephrase paragraphs for grammatical correctness and improved flow. We also used them to shorten text, making descriptions more concise and easier to read. All LLM-generated text was reviewed, edited, and approved by the human authors.

\section{Reproducibility, Release, and Intended Use}
\subsection{Reproducibility and release}
To support reproducibility, we will release (i) the \textsc{DeepFact-Bench} dataset, including de-identified annotations and claim metadata, and (ii) the \textsc{DeepFact-Eval} verifier code, prompts, and evaluation scripts. The code will be released under the \textbf{Apache-2.0} (or \textbf{MIT}) license, and the dataset under \textbf{CC BY 4.0} (or \textbf{CC BY-NC 4.0}) for research use. Where examples originate from third-party sources, we will follow their terms and, when necessary, distribute only derived metadata/identifiers rather than full text.

\subsection{Intended use and consistency with upstream terms.}
We use existing datasets and tools strictly for their intended research purpose: evaluating factuality and evidence-grounded verification, consistent with the licenses and access conditions specified by their authors. Our released artifacts---\textsc{DeepFact-Bench} and \textsc{DeepFact-Eval}---are intended for \emph{research-only} use in benchmarking and developing claim-level verifiers for Deep Research Reports (DRRs), including evaluation, ablations, and error analysis. To remain compatible with upstream access conditions, we avoid redistributing restricted third-party content when applicable and release only derived, de-identified annotations and metadata (e.g., claim text, verdicts, rationales, and provenance pointers) needed to reproduce our experiments. We explicitly prohibit non-research use that would violate upstream terms (e.g., commercial redistribution of restricted content or attempts to re-identify participants) and require users to comply with the original licenses/ToS of any upstream resources and retrieval services used in our pipeline.

\section{Qualitative Examples}

\subsection{Adversarial Examples}
\label{appendix:adversarial_examples}
Here we show examples of how we construct adversarial examples with intentional errors. We show one example each for collection error, analysis error, generalization error. 

\begin{tcolorbox}[colback=gray!5!white, colframe=gray!75!black, title=\textbf{Example: Collection Error}]
    \small
    \textbf{Context:}\\
    ...Intrinsic resistance genes, naturally encoded in soil microbial genomes, are primarily disseminated through vertical gene transfer (VGT) and limited horizontal gene transfer (HGT) under stable conditions. Acquired resistance genes, introduced via anthropogenic inputs like swine manure, rely heavily on MGEs (plasmids, transposons, integrons) to facilitate rapid HGT. Environmental factors such as soil moisture, pH, and heavy metals differentially impact these gene types, with acquired ARGs showing greater responsiveness to external stressors. The study by \textbf{Guo et al.\ (2025)} highlights the role of transposons like \emph{ISRj1} and \emph{IS91} in amplifying acquired resistance, while \textbf{Forsberg et al.\ (2012)} note that intrinsic resistance genes in soil producers (e.g., \emph{Streptomyces}) are less mobile but can persist for decades. Key findings reveal that swine farm soils exhibit higher acquired ARG diversity and MGE abundance compared to undisturbed soils, with winter conditions paradoxically enhancing their persistence. The report concludes that acquired resistance genes pose a greater risk for dissemination due to their mobility and environmental adaptability, while intrinsic genes remain a baseline reservoir. Limitations include the need for longitudinal studies and the challenge of distinguishing intrinsic from acquired genes in complex soil metagenomes. ...

    \tcbline

    \textbf{Original Sentence:}\\
    The study by \textbf{Guo et al.\ (2025)} highlights the role of transposons like \emph{ISRj1} and \emph{IS91} in amplifying acquired resistance, while \textbf{Forsberg et al.\ (2012)} note that intrinsic resistance genes in soil producers (e.g., \emph{Streptomyces}) are less mobile but can persist for decades.

    \tcbline

    \textbf{Adversarial Sentence:}\\
    The study by \textbf{Forsberg et al.\ (2012)} highlights the role of transposons like \emph{ISRj1} and \emph{IS91} in amplifying acquired resistance, while \textbf{Guo et al.\ (2025)} note that intrinsic resistance genes in soil producers (e.g., \emph{Streptomyces}) are less mobile but can persist for decades.

    \tcbline

    \textbf{Analysis:}\\
    The adversarial sentence performs an \emph{attribution swap}: it flips which paper supports each claim, subtly misassigning the transposon-driven amplification result (from Guo et al., 2025) and the intrinsic-gene mobility/persistence observation (from Forsberg et al., 2012). Because both papers are plausibly related, the swap can appear credible while corrupting provenance and authority.
\end{tcolorbox}

\begin{tcolorbox}[colback=gray!5!white, colframe=gray!75!black, title=\textbf{Example: Analysis Error}]
    \small
    \textbf{Context:}\\
    ...Studies over the past 30 years have consistently found tetracycline and sulfonamide antibiotic resistance genes (ARGs) to be ubiquitous in agricultural soils, even in the absence of recent antibiotic inputs
    \url{https://digitalcommons.unl.edu/agronomyfacpub/1098/}
    \url{https://www.frontiersin.org/journals/microbiology/articles/10.3389/fmicb.2018.01283/full}
    These genes are typically quantified in terms of gene copies per gram of soil, or as a ratio to total bacterial 16S rRNA gene copies. Across diverse farm soil environments, \emph{the abundance of tetracycline and sulfonamide ARGs generally falls in the range of $10^{4}$--$10^{6}$ gene copies per gram of soil}
    \url{https://pmc.ncbi.nlm.nih.gov/articles/PMC12031239/}
    \href{https://www.researchgate.net/publication/329356960_Fate_of_tetracycline_and_sulfonamide_resistance_genes_in_a_grassland_soil_amended_with_different_organic_fertilizers#:~:text=biosolid,}{\texttt{researchgate.net}}.
    \textbf{This corresponds to roughly $10^{-5}$--$10^{-3}$ ARG copies per bacterial 16S gene -- about one ARG per 1{,}000 bacterial cells in typical human-impacted soils}
    \url{https://ouci.dntb.gov.ua/en/works/lxYAxdL9/}
    Table 1 summarizes representative concentration ranges for common tetracycline (tet) and sulfonamide (sul) resistance genes reported in agricultural soils. \emph{Table 1. Typical abundance ranges of selected ARGs in agricultural soils (gene copies per g soil).} ...
    
    \tcbline

    \textbf{Original Sentence:}\\
    \textbf{This corresponds to roughly $10^{-5}$--$10^{-3}$ ARG copies per bacterial 16S gene -- about one ARG per 1{,}000 bacterial cells, in typical human-impacted soils}
    \url{https://ouci.dntb.gov.ua/en/works/lxYAxdL9/}
    
    \tcbline

    \textbf{Adversarial Sentence:}\\
    This corresponds to roughly 0.1--0.3\% of bacterial 16S genes carrying an ARG -- or one in every \textbf{300} bacterial cells, on average, in typical human-impacted soils.
    
    \tcbline
    
    \textbf{Analysis:}\\
    The adversarial sentence introduces a subtle quantitative distortion by changing units and the implied denominator: it reframes ``about one per 1{,}000 cells'' as ``one in every 300 cells'' while presenting the ratio as a percent. This inflates the perceived prevalence by roughly $\sim$3$\times$, which can be easily overlooked but materially changes the risk impression.
\end{tcolorbox}

\begin{tcolorbox}[colback=gray!5!white, colframe=gray!75!black, title=\textbf{Example: Generalization Error}]
    \small
    \textbf{Context:}\\
    ...It stands to reason that part of that predictive power would involve the listening sub-scores. Similarly, Sudina et al.\ (2021) assessed English proficiency holistically (with sections for listening and reading) and saw perseverance-linked gains
    \url{https://experts.nau.edu/en/publications/language-specific-grit-exploring-psychometric-properties-predicti}
    -- implying grittier students had stronger listening comprehension alongside other skills. One specific study in a mobile-assisted learning context evaluated beginners’ listening outcomes. In that study (Botes et al., 2025), 245 Duolingo learners’ listening and reading proficiency were tested; results indicated that \textbf{L2 grit was among the factors associated with better listening scores}, although motivation and age also played roles
    \url{https://www.researchgate.net/publication/391363862_L2_grit_and_age_as_predictors_of_attrition_in_mobile-assisted_language_learning}
    Furthermore, anecdotal evidence from EFL instructors suggests that gritty students are more likely to engage in extra listening practice (such as watching English media without subtitles, repeatedly listening to difficult audio until they understand, etc.), which over time enhances their listening ability. The limited direct research means we should be cautious, but no findings so far contradict the expectation that grit aids listening comprehension. If anything, listening might show a slightly weaker grit correlation than productive skills, simply because even a very persistent learner can still struggle with fast, unfamiliar speech. ...
    
    \tcbline

     \textbf{Original Sentence:}\\
    In that study (Botes et al., 2025), 245 Duolingo learners’ listening and reading proficiency were tested; results indicated that \textbf{L2 grit was among the factors associated with better listening scores}, although motivation and age also played roles
    \url{https://www.researchgate.net/publication/391363862_L2_grit_and_age_as_predictors_of_attrition_in_mobile-assisted_language_learning}

    \tcbline

    \textbf{Adversarial Sentence:}\\
    In that study (Botes et al., 2025), 245 Duolingo learners’ listening and reading proficiency were tested; results indicated that L2 grit was the \textbf{primary factor} determining listening scores, with little influence from motivation or age.

    \tcbline
    
    \textbf{Analysis:}\\
    The misleading sentence creates a \textit{conditional collapse} by removing caveats about motivation and age, incorrectly reframing grit as the sole or primary driver of listening outcomes and overstating the strength of the study’s conclusion.
\end{tcolorbox}

\subsection{Deep Research Errors Examples}
\label{appendix:deep_research_erros_examples}
Here, we show the errors deep research models make in the generated deep research reports. 

We identified 27.0\% naturally unsupported claims among the test set (excluding adversarially constructed examples).
These errors span several stages of the research pipeline—collection, analysis, and generalization—and reflect distinct reasoning failures, aligns with the taxonomy in \autoref{tab:error-taxonomy}:
Collection errors: fabricated claims without any identifiable source, or misattributions where a statement is linked to the wrong paper or dataset.
Analysis errors: misstatements of numerical results, incorrect mappings between experimental setups (e.g., attributing results from setup A to setup B), or faulty synthesis across sections (e.g., conflating Natural Questions with other multi-hop datasets).
Generalization errors: over-extending localized or conditional findings, such as extrapolating a global trend to a specific region without supporting evidence. Together, these categories highlight how unsupported claims in DRRs arise not only from missing citations but also from deeper reasoning and synthesis failures during evidence interpretation.

\begin{tcolorbox}[colback=gray!5!white, colframe=gray!75!black, title=\textbf{Example: Analysis Error}]
    \small
    \textbf{Context:}\\
    \textbf{5. Longitudinal and Multimodal Data Integration.}
    Longitudinal designs are underrepresented in L2 engagement research, with only 13.4\% of studies in the 2021 review employing them
    (\href{https://www.researchgate.net/publication/369536667_Engagement_in_language_learning_A_systematic_review_of_20_years_of_research_methods_and_definitions}{[1]}).
    A 2025 study on achievement emotions
    (\href{https://www.researchgate.net/publication/388364149_Do_achievement_emotions_underlie_L2_engagement_A_mixed-methods_multinational_study_on_the_role_of_achievement_emotions_in_multilingual_English_learners'_behavioral_cognitive_and_emotional_engagement}{[2]})
    used a mixed-methods explanatory sequential design (ESD) to link qualitative interview data with quantitative regression models.
    The study reports a strong association between achievement emotions and emotional engagement (r = .798, p = .000).
    Despite these insights, the cross-sectional nature of most studies limits their ability to capture engagement as a dynamic process.
    A 2021 behavioral analytics study
    (\href{https://www.engageli.com/blog/active-learning-statistics-2025}{[10]})
    proposed integrating multimodal data (e.g., facial expressions + keystroke dynamics) to address this gap, though its application to L2 contexts remains untested.
    \emph{Table 1: Comparison of Quantitative Methods in L2 Engagement Research.}

    \tcbline

    \textbf{Sentence:}\\
    The study found that positive emotions like hope and enjoyment correlated with higher engagement scores (r = 0.798), while negative emotions (e.g., anxiety) reduced participation.

    \tcbline

    \textbf{Human Verdict:}\\
    \textbf{Contradictory}

    \tcbline

    \textbf{Human Reason:}\\
    The strongest association was observed between achievement emotions and emotional engagement (r = .798, p = .000).
    However, the study did not distinguish between positive and negative emotions. Instead, achievement emotions are considered as a unified construct.
    See: \href{https://www.researchgate.net/publication/388364149_Do_achievement_emotions_underlie_L2_engagement_A_mixed-methods_multinational_study_on_the_role_of_achievement_emotions_in_multilingual_English_learners'_behavioral_cognitive_and_emotional_engagement}{source}.
\end{tcolorbox}

\begin{tcolorbox}[colback=gray!5!white, colframe=gray!75!black, title=\textbf{Example: Collection Error}]
    \small
    \textbf{Context:}\\
    \textbf{1. Cost and Insurance Coverage.} In the U.S., CAB-LA’s adoption is limited to 1.4\% of PrEP users due to insurance restrictions and high out-of-pocket costs
    (\href{https://pmc.ncbi.nlm.nih.gov/articles/PMC11776749/}{[3]}).
    Lenacapavir’s pricing (US\$28{,}000 per dose) further exacerbates access disparities
    (\href{https://pmc.ncbi.nlm.nih.gov/articles/PMC12178911/}{[10]}).
    \textbf{2. Regulatory Delays.} While CAB-LA is approved in the U.S. and Brazil, lenacapavir’s rollout in Europe and the Asia-Pacific is pending, with regulatory submissions in 2025
    (\href{https://www.prepwatch.org/tracking-lenacapavir-rollout/}{[11]}).
    \textbf{3. Healthcare Infrastructure.} In Zambia, inconsistent HIV testing protocols (e.g., 17\% RNA testing at first injection) may underrepresent seroconversions
    (\href{https://www.gilead.com/news/news-details/2025/gilead-presents-new-data-on-twice-yearly-lenacapavir-yeztugo-for-hiv-prevention-at-ias-2025}{[6]}).
    In the U.S., mental health and substance use comorbidities correlate with CAB-LA discontinuation
    (\href{https://www.who.int/news/item/14-07-2025-who-recommends-injectable-lenacapavir-for-hiv-prevention}{[4]}).
    \textbf{4. Stigma and Patient Preferences.} The \textbf{PILLAR trial} (2024) found that 75\% of U.S.\ participants preferred CAB-LA over daily oral PrEP, citing reduced stigma and convenience
    (\href{https://viivhealthcare.com/hiv-news-and-media/news/press-releases/2024/october/real-world-studies-for-apretude/}{[2]}).
    Similar data for lenacapavir is absent in the sources.

    \tcbline

    \textbf{Sentence:}\\
    The \textbf{PILLAR trial} (2024) found that 75\% of U.S.\ participants preferred CAB-LA over daily oral PrEP, citing reduced stigma and convenience. Similar data for lenacapavir is absent in the sources.

    \tcbline

    \textbf{Human Verdict:}\\
    \textbf{Contradictory}

    \tcbline

    \textbf{Human Reason:}\\
    The claim asserts that the PILLAR trial in 2024 found a 75\% preference for injectable cabotegravir (CAB-LA) over oral PrEP in the U.S., citing a ViiV Healthcare press release as its source. This is contradicted by multiple facts. First, the PILLAR study (NCT05422333) is an ongoing implementation study with an estimated completion date of December 2025, so it could not have produced final results in 2024
    (\href{https://clinicaltrials.gov/study/NCT05422333}{ClinicalTrials.gov}).
    Second, the cited ViiV press release discusses real-world effectiveness data from the OPERA and Trio Health cohorts; it does not mention the PILLAR trial, a 75\% preference rate, or reasons such as stigma and convenience
    (\href{https://viivhealthcare.com/hiv-news-and-media/news/press-releases/2024/february/viiv-healthcare-presents-first-of-its-kind-real-world-data-showing-the-high-effectiveness-of-apretude-cabotegravir-long-acting-for-prep-in-the-us/}{ViiV Healthcare}).
    While other studies (e.g., HPTN 083) have reported similar preference rates, this sentence’s specific attribution to the 2024 PILLAR trial and the provided source is factually incorrect.
\end{tcolorbox}

\begin{tcolorbox}[colback=gray!5!white, colframe=gray!75!black, title=\textbf{Example: Overclaim (Field-wide ``Most Promising'')}] 
    \small
    \textbf{Context:}\\
    For larger instances, its strength shifts to its sophisticated feasibility reasoning, making it an invaluable component within hybrid solvers. Its main weakness is its limited scalability when used as a pure, standalone optimization engine. \textbf{Reinforcement Learning} represents a paradigm shift from offline problem-solving to online policy learning. Its capacity to learn generalizable, adaptive scheduling policies that can be executed in near-real-time positions it as the most promising approach for dynamic and stochastic environments. However, it faces the highest implementation barriers, significant computational costs for training, and challenges in interpretability and achieving the absolute best solution quality compared to finely-tuned search methods. Ultimately, the most compelling future for academic research and practical application in construction project scheduling lies at the intersection of these three paradigms. The integration of RL's adaptive learning, CP's logical reasoning, and metaheuristics' powerful search capabilities promises to yield a new generation of hybrid, intelligent scheduling systems.

    \tcbline

    \textbf{Sentence:}\\
    Its capacity to learn generalizable, adaptive scheduling policies that can be executed in near-real-time positions it as the most promising approach for dynamic and stochastic environments.

    \tcbline

    \textbf{Human Verdict:}\\
    \textbf{Inconclusive}

    \tcbline

    \textbf{Human Reason:}\\
    Based on a review of recent (2022--2024) systematic surveys, no authoritative, up-to-date evidence establishes Reinforcement Learning (RL) as \emph{the most promising approach} for project scheduling in dynamic and stochastic environments compared to Metaheuristics and Constraint Programming. Recent reviews (e.g., Khajesaeedi et al., 2024; Bahroun et al., 2024) note RL strengths (generalizability, adaptivity, near-real-time inference post-training), but metaheuristics and hybrids remain widely recognized as state-of-the-art in research and practice. RL is often described as promising, yet there is limited systematic head-to-head evidence supporting overall superiority. Thus, the ``most promising'' claim is currently unsupported by field-wide consensus, yielding an evidence-based verdict of \textbf{INCONCLUSIVE}.
\end{tcolorbox}

\subsection{DeepFact-Eval Examples}
\label{appendix:deepfact-eval_examples}
Here we show representative cases where DeepFact-Eval succeeds and fails (Model output is simplified to be easy to read). DeepFact-Eval can decompose a sentence into atomic claims, cross-check the broader literature, and synthesize evidence to verify each claim. However, it can still err—for example, it may  miss critical evidence due to incomplete retrieval, retrieve closely matching evidence but miss key nuances or misinterpret it, or fail to validate niche sub-claims embedded within a longer sentence, suggesting room for improvements.

\begin{tcolorbox}[colback=gray!5!white, colframe=gray!75!black, title=\textbf{Example: DeepFact-Eval Success}]
    \small
    \textbf{Context:}\\
    When reference answers are available (e.g., a human-written answer for the question), automated metrics like \textbf{BLEU} or \textbf{ROUGE} have been used to measure n-gram overlap between the model’s answer and the reference. For instance, ROUGE-L was used in early LFQA benchmarks like ELI5
    (\href{https://aclanthology.org/2021.naacl-main.393/}{Krishna et al., 2021}).
    However, researchers found that such overlap metrics are \textbf{not very informative for long answers}---they can be gamed and do not truly reflect answer quality
    (\href{https://aclanthology.org/2021.naacl-main.393/}{Krishna et al., 2021}).
    Exact-match or F1 score (common in short-answer QA) become less applicable as answers grow longer and more free-form. More recently, embedding-based metrics (e.g., \textbf{BERTScore}) and Q\&A-based evaluation have been explored to capture semantic similarity. Yet a 2023 study reported that \textbf{no existing automatic metric reliably predicts human judgments} of long-form answer quality
    (\href{https://arxiv.org/abs/2305.18201}{Xu et al., 2023}).

    \tcbline

    \textbf{Sentence:}\\
    Exact-match or F1 score (common in short-answer QA) become less applicable as answers grow longer and more free-form.

    \tcbline

    \textbf{Human Verdict:}\\
    \textbf{Supported}

    \tcbline

    \textbf{Model Verdict:}\\
    \textbf{Supported}

    \tcbline

    \textbf{Model Reason:}\\
    The claim has two components: (i) EM/F1 are standard for short-answer QA, and (ii) they are less suitable for long, free-form answers.
    Evidence supports both:
    \begin{itemize}
        \item \textbf{Standard in short-answer QA:} EM/F1 are widely used in extraction-based / factoid QA evaluations (e.g., \href{https://www.sciencedirect.com/science/article/pii/S2949719124000360}{Upadhyay et al., 2024}; \href{https://arxiv.org/pdf/2108.06130}{Risch et al., 2021}).
        \item \textbf{Less applicable for long-form answers:} Overlap-style metrics correlate weakly with human judgments for long-form, open-ended QA and fail to capture semantic correctness and diversity of valid phrasings (e.g., \href{https://arxiv.org/abs/2305.18201}{Xu et al., 2023}; \href{https://aclanthology.org/2021.naacl-main.393/}{Krishna et al., 2021}; \href{https://arxiv.org/html/2407.04069v2}{Laskar et al., 2024}).
    \end{itemize}
    Therefore, the sentence is supported by multiple independent sources.
\end{tcolorbox}

\begin{tcolorbox}[colback=gray!5!white, colframe=gray!75!black, title=\textbf{Example: DeepFact-Eval finds closely-match citation but with nuances}]
    \small
    \textbf{Context:}\\
    For example, using a powerful model (like GPT-4) to directly score or rank a report on various criteria has become common. These \emph{LLM-as-a-judge} approaches can be more sensitive to nuances of content quality than static metrics. However, they come with reliability issues (biases and lack of transparency, discussed more below). As a workaround, some research proposes training dedicated \textbf{reward models} specialized for long-text evaluation
    (\href{https://arxiv.org/html/2502.19103v1}{Wu et al., 2025}).
    Such a reward model would be taught to predict human preferences on long outputs, similar to how RLHF trains a model to judge responses. This is an active research area: for instance, one 2024 work designed a reward model specifically to evaluate long academic essays, as standard metrics were insufficient
    (\href{https://arxiv.org/html/2502.19103v1}{Wu et al., 2025}).

    \tcbline

    \textbf{Sentence:}\\
    As a workaround, some research proposes training dedicated \textbf{reward models} specialized for long-text evaluation
    (\href{https://arxiv.org/html/2502.19103v1}{source}).

    \tcbline

    \textbf{Human Verdict:}\\
    \textbf{Contradictory}

    \tcbline

    \textbf{Model Verdict:}\\
    \textbf{Supported}

    \tcbline

    \textbf{Model Reason:}\\
    The cited LongEval paper states that, as an alternative to LLM-based scoring, prior work (e.g., Zhang et al., 2024c) proposes a reward model specifically designed for long-text evaluation, which directly supports the sentence
    (\href{https://arxiv.org/html/2502.19103v1}{Wu et al., 2025}).

    \tcbline

    \textbf{Human Reason:}\\
    While Wu et al., 2025 contains wording like ``Zhang et al., 2024c propose a reward model,'' the underlying referenced work does \emph{not} actually train a dedicated reward model; instead, it relies on existing LLMs as judges.
\end{tcolorbox}

\begin{tcolorbox}[colback=gray!5!white, colframe=gray!75!black, title=\textbf{Example: DeepFact-Eval ignores niche details}]
    \small
    \textbf{Context:}\\
    \textbf{Key Benchmarks.} \emph{No single standard benchmark yet.}
    \textbf{LONG2RAG (2024)} evaluates long-form answer quality and source usage
    (\href{https://aclanthology.org/2024.findings-emnlp.279.pdf}{source}).
    \textbf{Academic/Domain evals} (e.g., medical long-form QA with doctor evaluations)
    (\href{https://openreview.net/forum?id=8Qba6OeW9a}{source}).
    \textbf{Research agent tests}---e.g., \textbf{DeepResearcher (2025)} uses open-domain QA sets (NQ, HotpotQA, MuSiQue, etc.) as a testbed for research agents, evaluated with model-based judges
    (\href{https://arxiv.org/pdf/2504.03160}{source}).
    \textbf{Long-form generation benchmarks} (HelloBench, LongEval 2024) evaluate general long outputs
    (\href{https://arxiv.org/html/2502.19103v1}{source}).

    \tcbline

    \textbf{Sentence:}\\
    \textbf{Research agent tests} -- e.g.\ DeepResearcher (2025) uses open-domain multi-hop QA sets (NQ, HotpotQA, MuSiQue, etc.) as testbed for research agents
    (\href{https://arxiv.org/pdf/2504.03160}{source}),
    evaluated with model-based judges.

    \tcbline

    \textbf{Human Verdict:}\\
    \textbf{Contradictory}

    \tcbline

    \textbf{Model Verdict:}\\
    \textbf{Supported}

    \tcbline

    \textbf{Model Reason:}\\
    The DeepResearcher paper evaluates on multiple open-domain QA datasets including NaturalQuestions (NQ), HotpotQA, and MuSiQue, and uses a model-based judge (LLM-as-a-judge) to assess answer correctness rather than relying only on string-overlap metrics
    (\href{https://arxiv.org/pdf/2504.03160}{DeepResearcher, 2025}).

    \tcbline

    \textbf{Human Reason:}\\
    While DeepResearcher indeed evaluates on NQ, HotpotQA, and MuSiQue and uses model-based judging, the sentence characterizes \emph{NQ} as a \emph{multi-hop} QA set. Natural Questions is generally treated as \emph{single-hop} (factoid/short-answer style) rather than a multi-hop benchmark, so the sentence’s taxonomy claim is incorrect, yielding \textbf{CONTRADICTORY}.
\end{tcolorbox}

\end{document}